\documentclass[letterpaper,10pt,conference]{ieeeconf}
\IEEEoverridecommandlockouts
\usepackage{cite}
\usepackage{amsmath,amssymb,amsfonts}
\usepackage{algorithmic}
\usepackage{graphics} % for pdf, bitmapped graphics files
\usepackage{graphicx}
\usepackage{xcolor}
\usepackage[ruled]{algorithm2e}

\usepackage{tikz}
\usetikzlibrary{calc,positioning,fit} 

\usepackage{newtxtext}       % 
\usepackage{newtxmath}       % selects Times Roman as basic font
\usepackage{mathtools}
\usepackage{nicefrac}
\usepackage{balance}
\usepackage{subcaption}
\usepackage{graphicx}
\usepackage{amsmath}
\usepackage{multirow}
\usepackage[]{algorithm2e}
\usepackage[strings]{underscore}

\usepackage{changepage}

\newcommand{\algName}{{\em NavTuner}}

\makeatletter
\def\endthebibliography{%
	\def\@noitemerr{\@latex@warning{Empty `thebibliography' environment}}%
	\endlist
}
\makeatother

\def\BibTeX{{\rm B\kern-.05em{\sc i\kern-.025em b}\kern-.08em
    T\kern-.1667em\lower.7ex\hbox{E}\kern-.125emX}}
\begin{document}

\title{\LARGE \bf NavTuner: Learning a Scene-Sensitive Family of Navigation Policies}

\author{Haoxin Ma$^{1}$, Justin S. Smith$^{2}$, and Patricio A. Vela$^{2}$% <-this % stops a space
\thanks{*This work supported in part by NSF Award \#1849333.}% <-this % stops a space
\thanks{$^{1}$H. Ma is with College of Computing, Georgia Institute of Technology, Atlanta, GA 30308, USA.
		{\tt\small haoxin.m@gatech.edu}}
\thanks{$^{2}$J.S. Smith, and P.A. Vela are with the School of Electrical and Computer Engineering and the Institute for Robotics and Intelligent Machines, Georgia Institute of Technology, Atlanta, GA 30308, USA.
        {\tt\small \{jssmith, pvela\}@gatech.edu}}%
}

\newcommand{\topSpace}{\vspace*{0.06in}}

\maketitle
\thispagestyle{empty}
\pagestyle{empty}

\begin{abstract}
  The advent of deep learning has inspired research into end-to-end
  learning for a variety of problem domains in robotics.
  For navigation, the resulting methods may not have the generalization
  properties desired let alone match the performance of traditional
  methods.
  Instead of learning a navigation policy, we explore learning an
  adaptive policy in the parameter space of an existing navigation
  module.
  Having adaptive parameters provides the navigation module with a
  family of policies that can be dynamically reconfigured based on the
  local scene structure, and addresses the common assertion in machine
  learning that engineered solutions are inflexible.
  Of the methods tested, reinforcement learning (RL) is shown to provide
  a significant performance boost to a modern navigation method through
  reduced sensitivity of its success rate to environmental clutter.  The
  outcomes indicate that RL as a meta-policy learner, or dynamic
  parameter tuner, effectively robustifies algorithms sensitive to
  external, measurable nuisance factors.
\end{abstract}

\section{Introduction \label{sIntro}}
%Advances in the computer vision based on methods from machine learning
%and deep learning have translated to improved perception modules for a
%variety of robotics challenges, including 
%manipulation \cite{COUPLE},
%autonomous vehicles \cite{COUPLE}, and 
%\textcolor{red}{OTHER THEMES? maybe UAV?} \cite{COUPLEOTHER}.  
%Regarding autonomous vehicles, most of the advances are for structured
%navigation settings where visual markers serve to guide the vehicle
%\cite{SIGNDETECT,DeepDriving,ETC}.  
%For the less structured aspects of navigation, such as vehicles,
%pedestrians, and unforeseen road obstacles, learning could still improve
%\cite{SURVEYPAPERMAYBE}.
%The situation degrades when considering autonomous operation in
%even less structured environments, such as parking lots, off-road
%environments, or unmarked man-made environments. Yet, we anticipate
%automous operation of mobile robots. This paper explores how learning
%may improve decision making for collision avoidance in unstructured
%environments.

Autonomous navigation through static, unstructured environments has
advanced in the past decades but fundamentally still relies on engineered
approaches \cite{TEB,smith2020egoteb}.  Given an approximate map, the
approaches use sensor data to inform updated estimates of the
environment which are used to evaluate future trajectories in terms of
safety and other characteristics, with the aim of finding a
collision-free, goal-attaining path. Traditionally designed systems
involve manual parameter selection for general purpose navigation, which
exhibits sensitivity to environmental conditions.

This paper investigates the use of machine learning to dynamically
reconfigure the parameters of a hierarchical navigation system according
to the immediate, sensed surroundings of the robot.  We show that
scene-dependent online tuning improves navigation performance and
reduces sensitivity to environmental conditions.  The final
reinforcement learning solution, called {\algName}, addresses the
problem of parameter sensitivity to operational variance.

%The final reinforcement learning solution, called {\algName}, provides a
%means to improve the performance and robustness of traditionally
%engineered navigation systems to operational variance.

\subsection{Research Context\label{sRelated}\label{sResCon}}
\begin{figure*}[t]
  \topSpace
	\begin{subfigure}{.24\textwidth}
		\centering
		\includegraphics[width=1.30in]{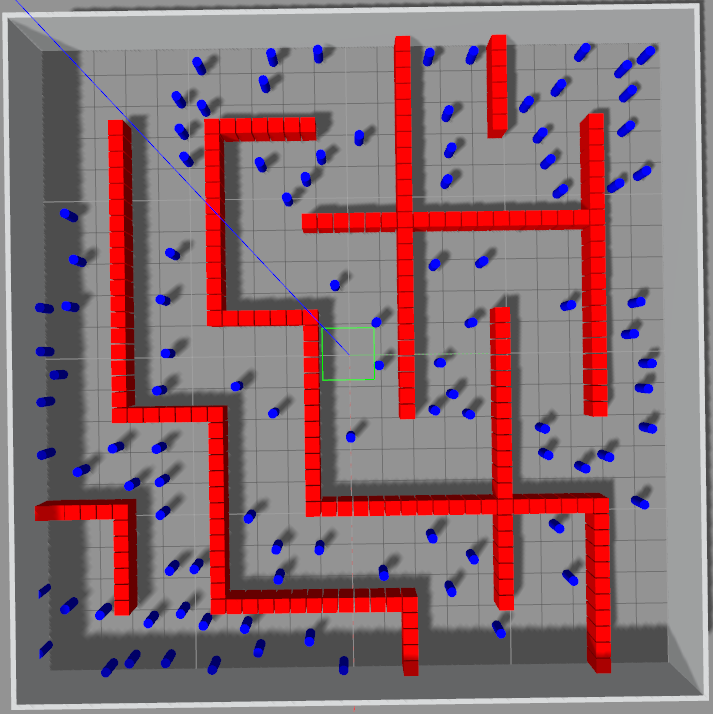}
	\end{subfigure}
	\begin{subfigure}{.24\textwidth}
		\centering
		\includegraphics[width=1.30in]{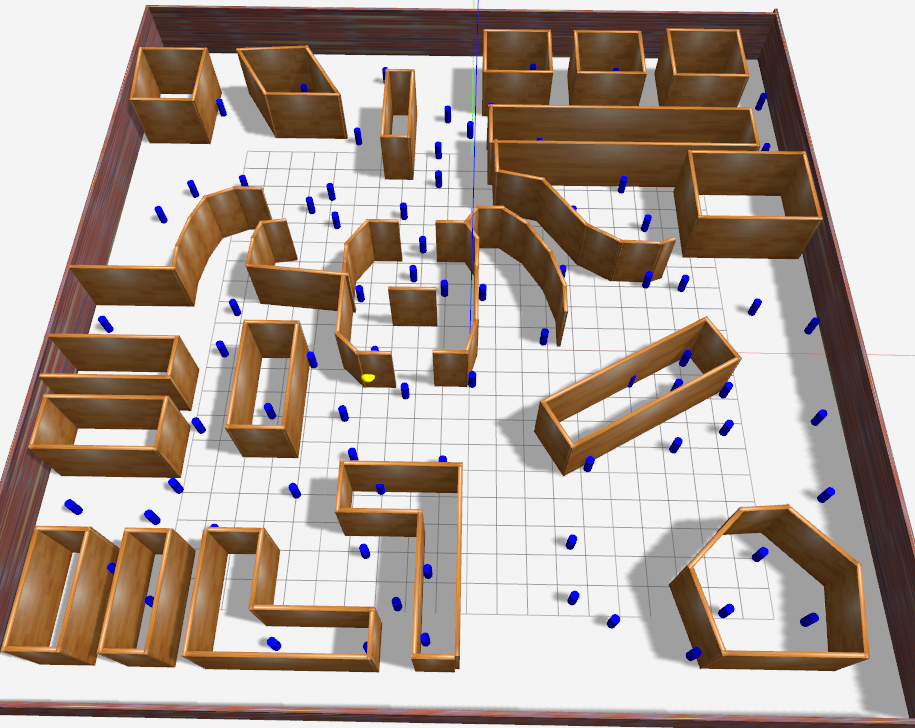}
	\end{subfigure}
	\begin{subfigure}{.24\textwidth}
		\centering
		\includegraphics[width=1.30in]{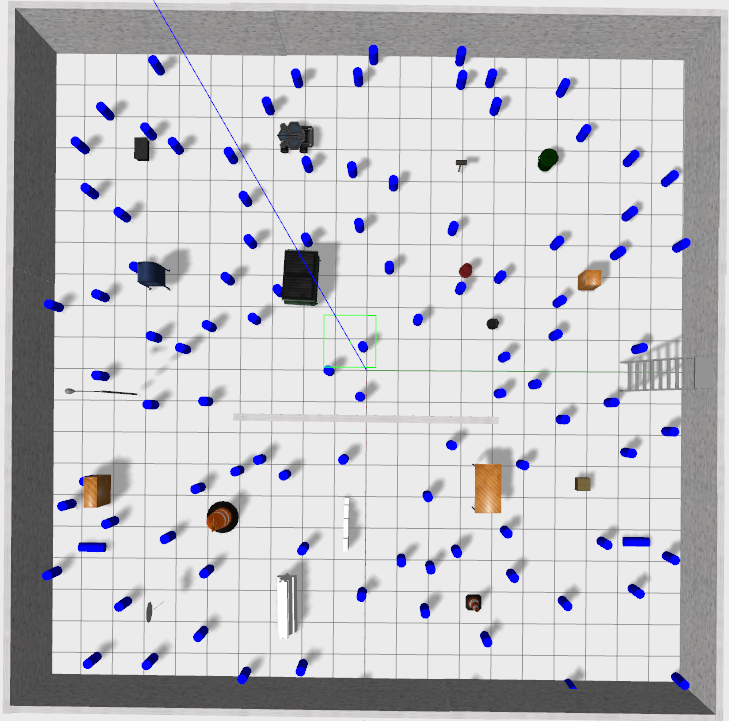}
	\end{subfigure}
	\begin{subfigure}{.24\textwidth}
		\centering
		\includegraphics[width=1.30in]{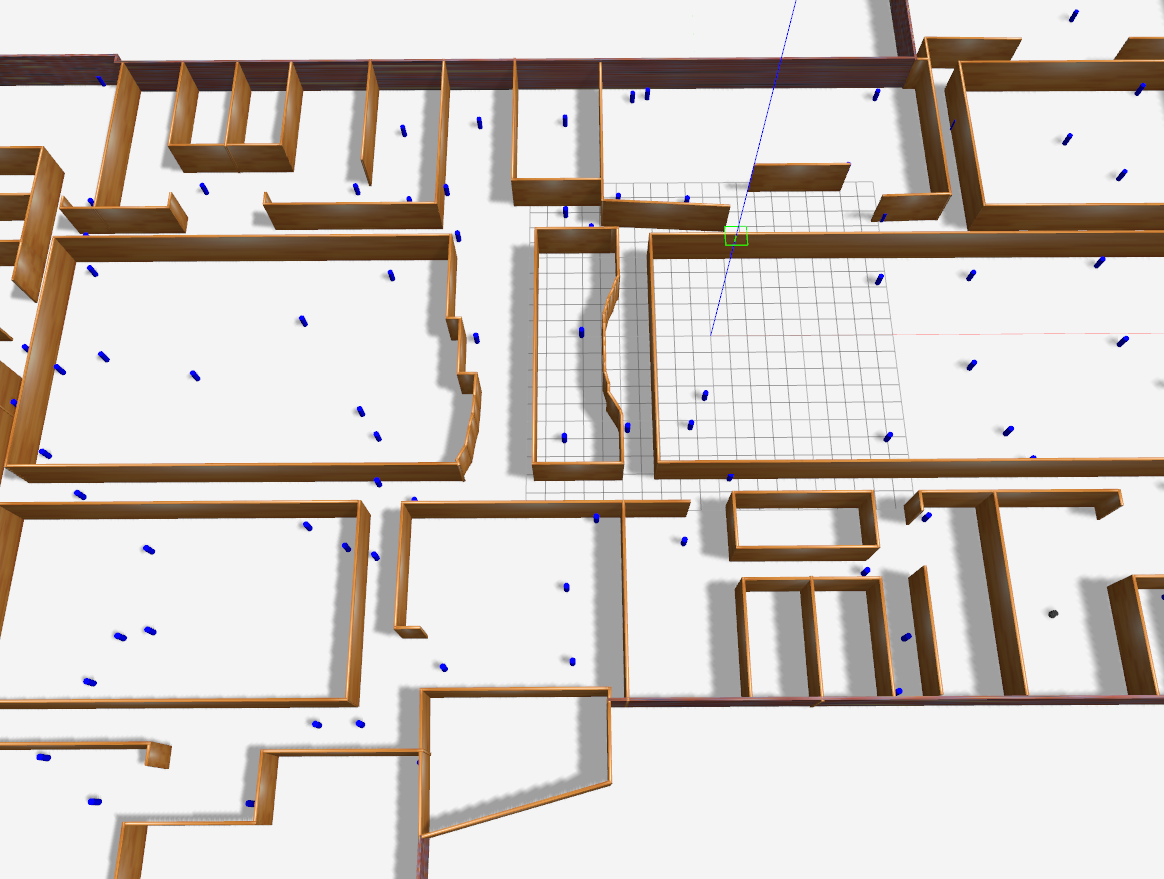}
	\end{subfigure}
	\begin{subfigure}{.245\textwidth}
		\centering
		\includegraphics[width=\linewidth]{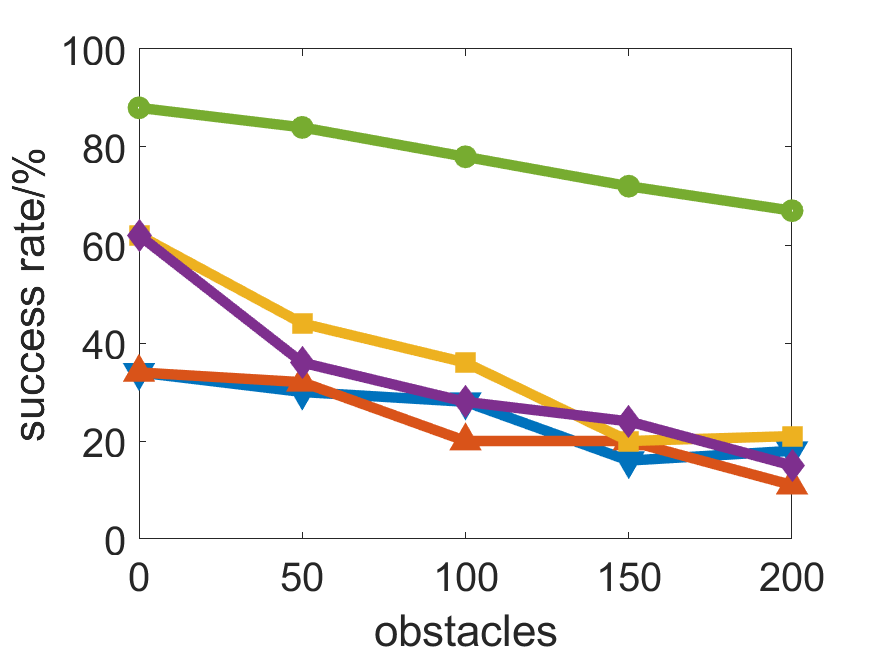}
	\end{subfigure}
	\begin{subfigure}{.245\textwidth}
		\centering
		\includegraphics[width=\linewidth]{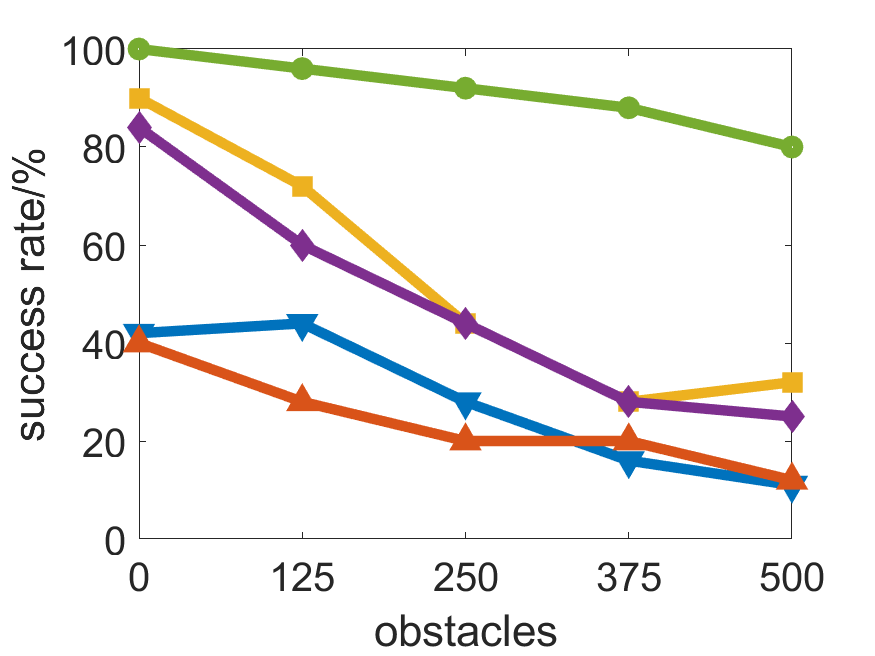}
	\end{subfigure}
	\begin{subfigure}{.245\textwidth}
		\centering
		\includegraphics[width=\linewidth]{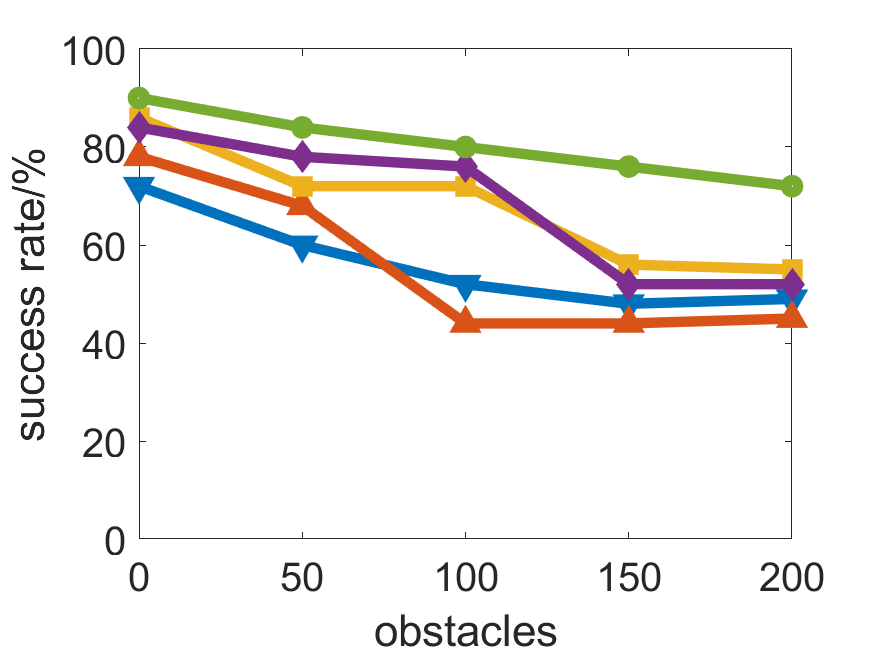}
	\end{subfigure}
	\begin{subfigure}{.245\textwidth}
		\centering
		\includegraphics[width=\linewidth]{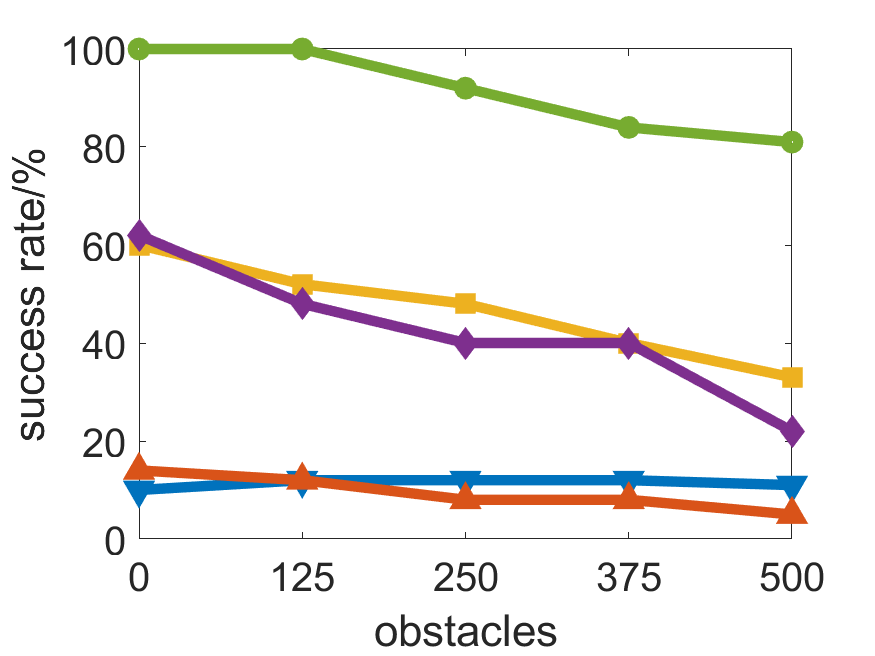}
	\end{subfigure}
	\centering
	\begin{subfigure}{.75\textwidth}
		\centering
		\includegraphics[height=1em]{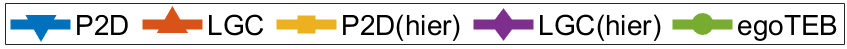}
	\end{subfigure}
	\caption{Benchmark Worlds their Results. Top row: Depictions of the
    environments Maze, Campus, Sector, and Office with 100 obstacles.
    Bottom row: Navigation success rate versus obstacle count plots for
    egoTEB \cite{smith2020egoteb} and methods from
    \cite{SmEtAl_Drive[2018],PerceptionToDecision}.\label{fig:benchmark}}
    \vspace*{-1.0em}
\end{figure*}

\subsubsection{Navigation and Machine Learning\label{NavML}}
One candidate approach to learning and navigation is to replace the
traditionally engineered system with an end-to-end sensor to decision
neural network
\cite{SmEtAl_Drive[2018],intention-net,PerceptionToDecision,XiEtAl[2020]NavLearnSurvey}. 
Empirical and limited benchmarking show some promise on this front.
However, instead of directly solving the navigation problem itself,
these methods solve some highly specific subset of it, typically
equivalent to the local navigation problem.  All argue for the need to
integrate with existing hierarchical schemes, though few actually do so.
Both \cite{SmEtAl_Drive[2018],PRM-RL} show that reinforcement learning
or reward-based approaches for local planning in a local-global
hierarchical planner can work.
Instead of mapping sensor input to navigation decisions directly, other
learning based methods seek to map the sensor data to higher level
information for decision making.
Inverse reinforcement learning has been used to learn a reward function
from RGB-D features and goal-directed trajectories \cite{Kim2016}. 
%Also, it's possible to take advantage of RL in a local-global navigation
%stack, for example, using probability roadmaps (PRM) for global planning
%and RL for local planning with laserscan and relative polar goal as
%input\cite{PRM-RL}. 
As has learning a neural SLAM model to output necessary information
for global and local policies to control a robot \cite{chaplot2020learning}.
Learning-based perception can be combined with model-based control
methods to navigate in partially observable, unknown environments
\cite{bansal2020combining}.

Overall, there is no substantive benchmarking of learning based methods
with traditional navigation schemes \cite{XiEtAl[2020]NavLearnSurvey}. 
Thus, the assertions that learning can overcome sensitivity to
environmental conditions and can outperform traditionally engineered
systems remains unconfirmed.  As a preliminary investigation, we
implemented a couple methods and benchmarked them in simulated
ROS/Gazebo environments, see Figure \ref{fig:benchmark} (and
\S\ref{evalE2E} for more details).  When compared to a traditional
hierarchical navigation approach \cite{smith2020egoteb}, learning-based
navigation methods have lower performance and equivalent or higher
sensitivity to the environment.  The sensitivity can be seen by the drop
in performance (e.g., higher slope) as the environment becomes denser in
the success rate vs obstacle quantity graphs of the second row.  Lastly,
the variable performance across environments indicates poor
generalization to world structure by the learning methods.

\subsubsection{Reinforcement Learning-Image to Action}
Tasks involving perception to action pipelines, such as visual
navigation, can be implemented using end-to-end reinforcement learning
(RL) based on deep convolutional neural network policy learners 
\cite{self-supervised-rl-gen-graphs,target-driven-visual-rl, bruce2017one, DeepVisuoMotor, Choi2011}.
This is perhaps the most pervasive use of RL in vision-based navigation
contexts. The learnt policy directly maps visual sensor input
(laserscan, RGB image, or depth map) to actions, steering commands, or
velocities \cite{virtual-to-real-rl,target-driven-visual-rl,drl-successor-features}.  
Policy learning can be improved by using knowledge from previously learnt visual navigation tasks when learning
new tasks \cite{drl-successor-features}.  Adding decision relevant
auxiliary tasks can promote learning internal representations that support 
navigation \cite{nav-aux-tasks}, and improve the sample efficiency of RL while 
speeding up the training process\cite{ye2020auxiliary}.
In a similar vein, providing mid-level visual cues improves the
learning, generalization, and performance of policies \cite{sax2019learning}.
Also, learning subroutines instead of individual actions can boost the performance\cite{kumar2020learning}.
% NOT RELATED TO PROBLEM. EVEN THE PAPER STATES AS MUCH. 
% One means to address the compute demands is to distribute the training across a compute network \cite{wijmans2019dd}.  Given that imagery can suffer from distribution shift problems, more
While navigation is a good task to demonstrate RL methods, in many cases
the task focuses on navigation guidance more so than collision
avoidance.  Evaluation does not focus on embodied navigation nor
realistic motion and collision models.  
Application of RL to robotics places more emphasis on the structure of
the learning network \cite{virtual-to-real-rl,gao2017intention} and 
on the use of existing navigation methods to generate policy samples 
\cite{gao2017intention,PerceptionToDecision,8326229}, and less on the policy
learner. 

\subsubsection{Vision-Based Navigation}
Typical solutions for vision-based navigation rely on a combination of
path planning and sensor-based world modeling to synthesize local paths
through the world based on recovered structure. The most effective
policies are hierarchical, with a \textit{global planner} establishing
potentially feasible paths based on an estimated map and a \textit{local
planner} for following as closely as possible the global path, subject
to collision-avoidance constraints in response to sensed obstacles
missing from the map
\cite{dynamicwindow,teb-first,uneven-unstructured-indoor}.
Though exhibiting strong performance in Figure \ref{fig:benchmark}, the
variable maximum success rates and downward trends of egoTEB
\cite{smith2020egoteb} across the graphs indicates that traditional
methods are not immune to variation in world structure. 
There is potential value in modifying them based on the local obstacle
configuration space, i.e., in performing online parameter tuning for
navigation.

\subsubsection{(Hyper)-Parameter Tuning} 
The potentially negative impact of incorrect hyper-parameters or
parameters is well established in machine learning. 
Hyper-parameter optimization improves learning outcomes without
requiring major modifications to the underlying learning structure
\cite{feurer2019hyperparameter,yu2020hyper,huang2019automatic}. 
When feasible to implement, the additional computation needed is offset
by the performance enhancement.  Given that RL policies can exhibit high
variability to the training process and parameter settings, the process
of gradient-free, hyper-parameter and reward tuning is recommended when
the additional computational resources are available \cite{AutoRL}.  
%More broadly, adaptive control \tocite{X} and control-based RL
%\tocite{REVIEW} are online parameter tuning methods for ensuring
%consistent performance of a controller in the face of unknown parametric
%variation or unmodeled dynamics.  For complex systems, it is
%sub-optimal for the designer to manually specify parameter values or
%functional structure when there is significant variance available (i.e.,
%there is a non-trivial manifold of valid instances) and when
%performance-relevant knowledge is unavailable. 

There is prior work on automatic parameter tuning applied to motion planning 
and navigation.  Motion planning algorithms can perform better with
random or Bayesian based \cite{cano2018automatic}, and model based
\cite{burger2017automated} automatic parameter tuning algorithms. 
The benefits also apply to safety constraints \cite{berkenkamp2016bayesian}.
Parameter tuning can also be incorporated into the learning process of a
learning-based motion planning algorithm to reduce sensitivity to
manual tuning \cite{bhardwaj2020differentiable}. Adaptive planner
parameters of a planner can be learnt through intervention \cite{wang2020appli}, demonstration \cite{xiao2020appld}, 
and reinforcement learning \cite{xu2020applr}. 
%The learned adaptive parameters can outperform default parameters, and
%generalize well to unseen environments.

%
%\textit{no hyperparameter tuning found in navigation field, but AutoRL is similar, which searches for structures instead of hyper-param values. They use what they call "gradient-free hyperparameter optimization", but the hyper-param here is defining the network structure and agent reward, not exactly the same with our "hyper-parameter".}
%
%Chiang \etal\cite{AutoRL} combines deep RL with gradient-free hyperparameter optimization to tune the agent’s reward and network architectures.

\section{Scope of Investigation \label{sScope}}
\newcommand{\fGP}{f_{\text GP}}
\newcommand{\dLA}{d_{\text LA}}

This paper explores the performance impact of online tuning for
traditional navigation strategies.  Per Figure \ref{fig:benchmark} and
\cite{wang2020appli,xiao2020appld,xu2020applr}, they exhibit sensitivity
to external world structure. Some of the sensitivity is a function of
manually specified algorithm constants (i.e., parameters) that do not
generalize well to different free-space circumstances.  The objective is
to reduce system sensitivity and improve performance outcomes during
deployment, by learning a more optimal tuning or adaptive policy for the
navigation parameters as a function of local geometry.  The study here is
a bit more systematic and extensive in relation to comparable studies.

\subsection{Hierarchical Navigation}

The system to study is a hierarchical, vision-based navigation system 
consisting of two distinct spatial-and time-scales, {\em global} and 
{\em local}, with planning modules associated to each scale. The global
planner computes a candidate path connecting the robot's current pose to
the goal pose based on the current map.  The local planner uses the
global path to generate a series of sub-goals that should be feasible to
sequentially achieve.  Since the environment is unknown or uncertain,
the local planner solves the current sub-goal based on a local map
informed by the real-time integration of sensory information regarding
the structure of the local environment. Its spatial and temporal scale
is sufficiently small that real-time processing of the sensory data and
short-time trajectory synthesis with collision checking and trajectory
scoring is achievable.
%\textcolor{orange}{
%For our experiments, we use the NavFn from ROS as the global planner,
%and the egoTEB controller as the local planner.  The egoTEB controller
%is a local planner that combines ego-centric, perception space
%representations for local planning 
%with the Timed-Elastic-Bands algorithm \cite{keller2014planning}. 
%}
%planner. The global planner will compute a global path
%from the start to the goal, which consists of a series of sub-goals, 
%and the local planner will take in the sub-goals and compute the
%controlling commands that move the robot to the current sub-goal. 
During navigation, information flow between the two planners influences
the paths planned and the paths taken. Parameters internal to the global
and local planners further influence their outputs.
The parameters and the interacting modules should work to achieve
collision-free navigation to the goal point while taking the shortest
path possible given {\em a priori} known and {\em a posteriori} sensed
information about the world. Here, the reference hierarchical planning system 
will utilize the {\em egoTEB} local planner \cite{Smith2020}.

\subsection{Navigation Parameters}

To illustrate the sensitivity of navigation performance to internal
parameters, this section describes the impact of two:
the {\em global planning frequency} and 
the local planner {\em look ahead distance}.  

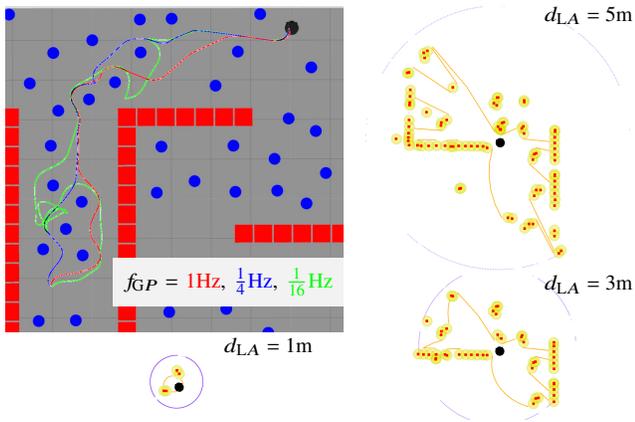
\begin{figure}[t]
  \topSpace
  \centering
  \begin{tikzpicture}[inner sep=0pt,outer sep=0pt]
    % Frequency and paths.
	\node (F) at (0,0) 
      {\includegraphics[width=0.52\columnwidth]{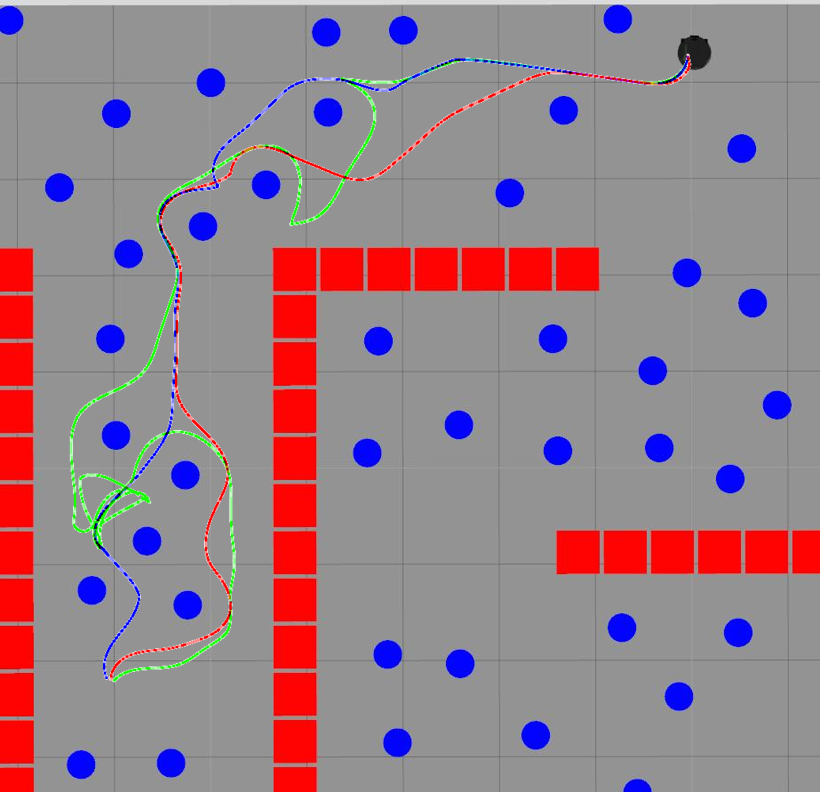}};

    % Look ahead distance.
    % dLA = 5.
    \node[anchor=north west,xshift=0.5em] (L5) at (F.north east)
	  {\includegraphics[width=0.425\columnwidth]{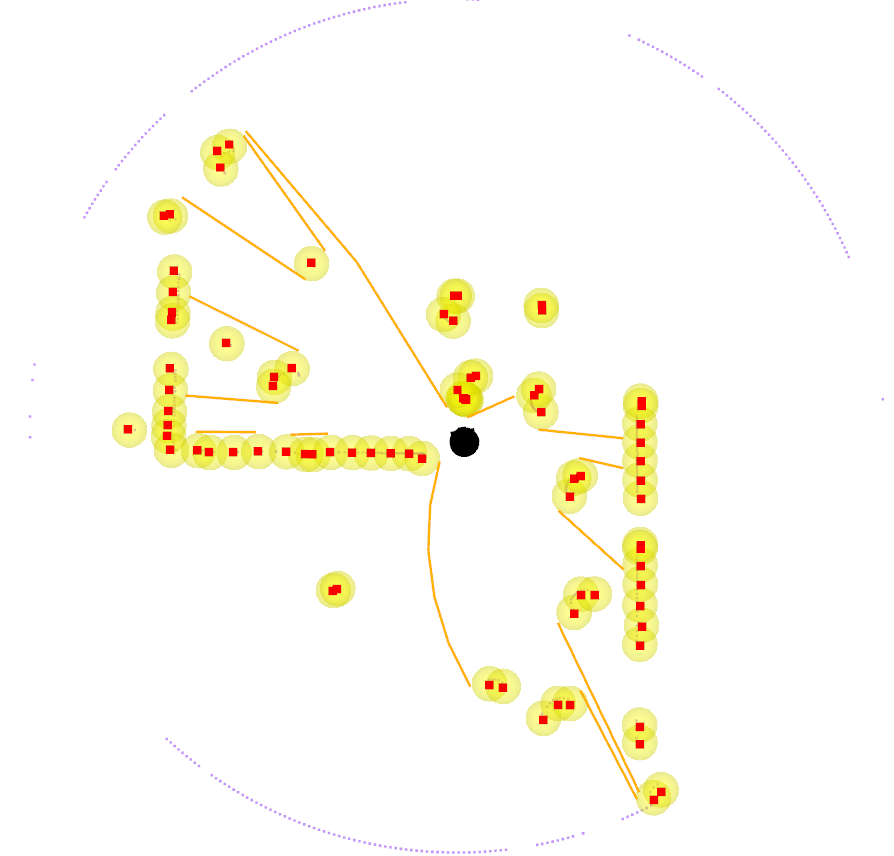}};
    % dLA = 3.
    \node[anchor=north,yshift=-2pt] (L3) at (L5.south)
	  {\includegraphics[width=0.425\columnwidth,clip=true,trim=0in 2.0in 0in 2.0in]{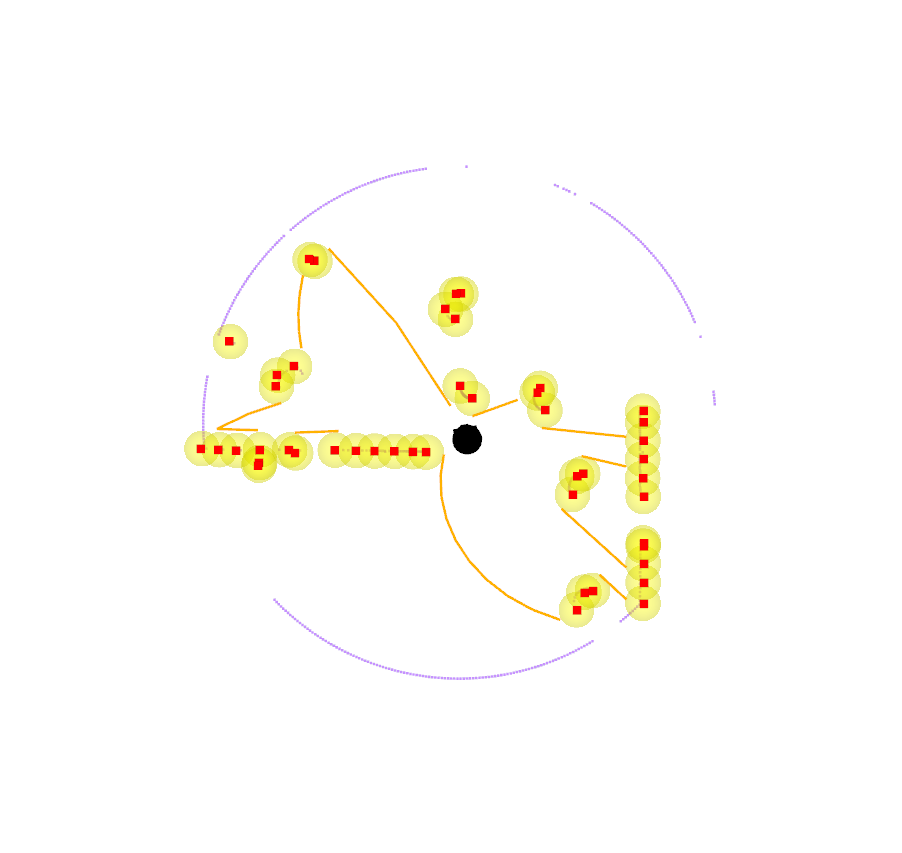}};
    % dLA = 1;
    \node[anchor=north,yshift=-2pt] (L1) at (F.south)
	  {{\includegraphics[width=0.425\columnwidth,clip=true,trim=0in 3in
      0in 3in]{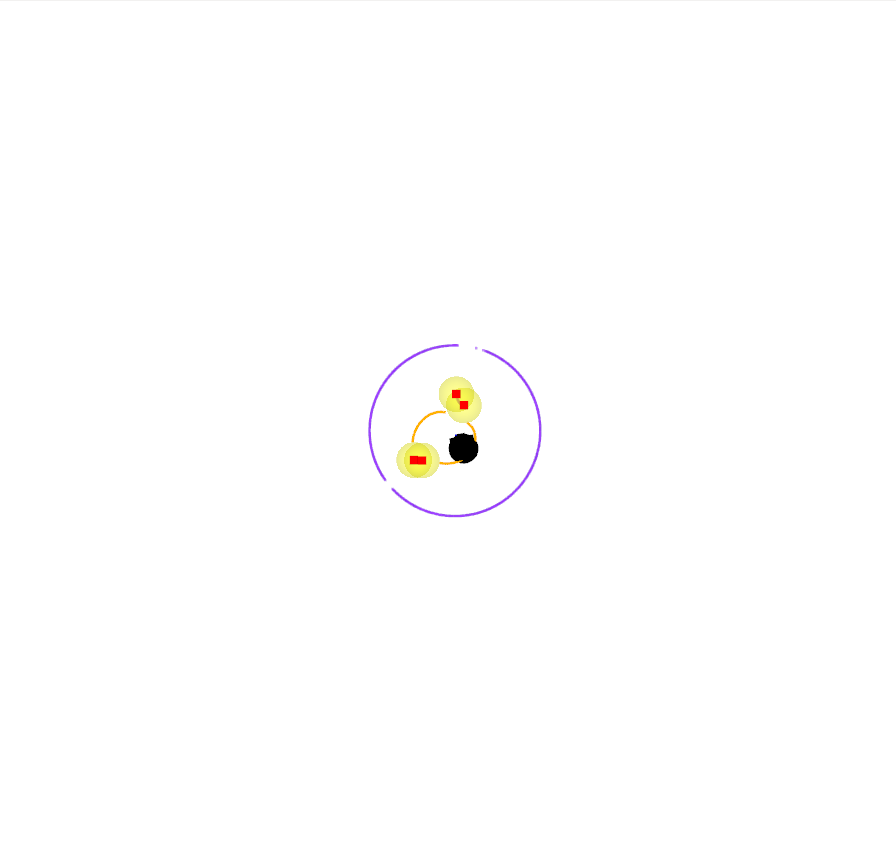}}};

    % Text labels.
    \node[anchor=north east] at (L5.north east) {\footnotesize $\dLA = 5$m};
    \node[anchor=north east] at (L3.north east) {\footnotesize $\dLA = 3$m};
    \node[anchor=north east] at (L1.north east) {\footnotesize $\dLA = 1$m};
    \node[anchor=south east,yshift=10pt,fill=white!90!gray,inner sep=4pt] 
      at (F.south east)
      {{\footnotesize \textcolor{black}{$\fGP =$} \textcolor{red}{$1$Hz}, 
                       \textcolor{blue}{$\frac 14$Hz}, 
                       \textcolor{green}{$\frac 1{16}$Hz} }};

  \end{tikzpicture}
  \caption{Changing planning frequency $\fGP$ impacts path taken (top-left).
  Changing look ahead distance $\dLA$ impacts the candidate navigable
  gaps found (orange curves).\label{impactFPG}\label{impactDLA}} 
  \vspace*{-0.5em}
\end{figure}

\subsubsection{Global Planning Frequency ($\fGP$)}
The global planner recomputes the global path at a specified frequency
to refresh its estimated best path based on data accumulated during
navigation.  Additionally, there are special events that trigger new
plans (such as arriving at a dead-end).  The tunable global planning
frequency is upper bounded by the local planning frequency (or sensing
rate) and the rate at which new global paths are generated.  Other
environmental events not part of the detect special events impact the
need for new global plans.  Sensory information accumulating in the
local map may indicate that the current global path is no longer
feasible and should be recomputed;  or that an unavailable path option
may now be feasible and should be considered.

The effect of changing $\fGP$ for a given environment and navigation
goal is visually exemplified in Figure \ref{impactFPG} (top-left). 
%A high frequency degrades navigation.
Table \ref{tab:diff} quantifies its influence. The table reports the
difference between the best performing and the worst performing outcomes
in the {\em Maze} environment as a function of $\fGP$ across the
different obstacle densities tested (50 repeated trials each).  
The difference is a measure of parameter sensitivity to environment
changes.  SR stands for success rate and PL stands for path length.  In
the environment with the most obstacles (least spacing) there are almost
20 more failures cases for the worst $\fGP$ setting versus the best.
The path length increases by 7.68m which is rougly 14\% of the average
path length.  

\begin{table}[t]
  \topSpace
	\centering
	\caption{Performance changes vs. inter-obstacle spacing.%
      \label{tab:diff}}
	\begin{tabular}{|c|c|c|c|c|}
		\hline
%		%average: (dLA)
%		%         0.75  1.0  1.25  1.5
%		%success  28.9 40.7  44.4  46.4
%		%path    26.69 30.97 29.62 29.31
%		%average: (fGP)
%		%         0.75  1.0  1.25  1.5
%		%success   26  39.6  43.2  46.6
%		%path    54.71 47.35 35.29 30.17
		\multirow{2}{*}{Spacing} & \multicolumn{2}{c|}{$\fGP$} &
        \multicolumn{2}{c|}{$\dLA$} \\\cline{2-5}
		& $\Delta$SR (\%) 
        & $\Delta$PL (m) 
        & $\Delta$SR (\%) & $\Delta$PL (m) \\\hline
		0.75 
            & 19 & 7.68
            & 8  & 8.65
        \\\hline
		1.0 
            & 8  & 1.56
            & 5  & 2.93
        \\\hline
		1.25 
            & 4  & 1.04
            & 7  & 2.34 
        \\\hline
		1.5 
            & 5  & 0.48
            & 6  & 1.8
        \\\hline
	\end{tabular}
\end{table}

\subsubsection{Look Ahead Distance ($\dLA$)}
EgoTEB relies on the egocircle, an egocentric polar obstacle data
structure, similar to a 1D laserscan, that operates as a local map. It
retains multiple measurements per scan entry and complements available
sensory information. For example, when the egocircle has a finer
resolution than the actual onboard sensor, it has obstacle readings not
captured by the sensor at the current time but captured at an earlier
time. This data structure is parsed by egoTEB to establish navigation
gaps between obstacles through which the robot may traverse.  The gaps
represent different path opportunities to take to the goal state. 

Gap processing uses $\dLA$ to define a look-ahead distance cut-off when
detecting gaps. 
Measurements beyond $\dLA$ are "ignored" in simulation of a shorter
distance scanner.  Local paths generated as part of egoTEB will not
extend beyond this radius. $\dLA$ limits the quantity of gaps detected
since only nearby obstacles are considered, which reduces the quantity
of candidate paths generated and tested.  Figure \ref{impactDLA} depicts
the local map information and the detected gaps based on three different
$\dLA$ values.  The parameter determines the spatial extent of the local
map, which ultimately influences several parts of the local navigation
strategy.  The impact of $\dLA$ on navigation is evident in Table
\ref{tab:diff}. 

%the local planner will look further ahead, and
%there will be more candidate trajectories to score, which means it will
%be less likely to find the optimal trajectory if there exist many
%possible trajectories. Therefore, we pick the egocircle radius $R$,
%which can be also referred to as the lookahead distance, as one of the
%hyper-parameters to vary. The other hyper-parameter we choose is global
%planning frequency, which determines how often the global planner
%refresh the global path. 

%The egocircle is able to populate, propagate, and store the local
%environmental history. In the Ego-TEB controller, the egocircle will be
%used to score candidate trajectories generated by the TEB local planner.
%The values at $\alpha$ of an egocircle $V_\alpha = \min(R_\alpha, R)$,
%where $R_\alpha$ stands for the range at $\alpha$ and $R$ is the
%egocircle radius, a hyper-parameter. The value of $R$ acts as a
%lookahead distance, as the egocircle will only include obstacles within
%the distance of $R$, and the local paths generated will also be within
%$R$. With larger $R$, 

\subsection{Scene Adaptive Policy Models}

Learning a dynamic tuning policy can be done through a variety of
mechanisms, with one main difference being batch learning versus
reinforcement learning (RL).  Both are explored and evaluated in this
study. Batch learning requires sampling the input and output
space to generate the training data. In contrast, RL involves rollouts
that test output values chosen according to some training policy and
record the measured input values to generate the training data.  The
input data is the egocircle reconstruction of the current environment
local to the robot. The output data will be the parameter choices.
%Given that $\fGP$ and $\dLA$ influence the performance of egoTEB
%navigation, the objective is to identify a tuning function or policy,
%e.g. a NavTuner, that can achieve better overall performance by tuning
%the two values during online operation. Here we describe the different
%learning structures and methods used to generate the NavTuner.
The discretization for $\dLA$ ranges over $[1.0\,,5.5]$ meters in $0.5$
increments (10 total), and for $\fGP$ ranges over $[\nicefrac1{16}\,, 1]$
Hz in factor of $2$ increments (5 total).

\subsubsection{Batch Learning\label{batchLearn}}
Several network models were chosen for the batch learning approach.
They include: a linear network, a neural network (w/ReLu), and a
convolutional neural network (w/ReLu).
Two output structures are tested for each, a classifier structure
over discretized values and a regression model for continuous
prediction (trained with the discretized values).  Training is performed
by collecting the performance statistics of constant parameter value
runs across a sweep of each discretized parameter individually. 
%with the other fixed. 

\subsubsection{Reinforcement Learning\label{parallel DQN}}
\label{sScopeRL}
\begin{figure}[t]
  \topSpace
  \centering
  \includegraphics[width=0.9\linewidth]{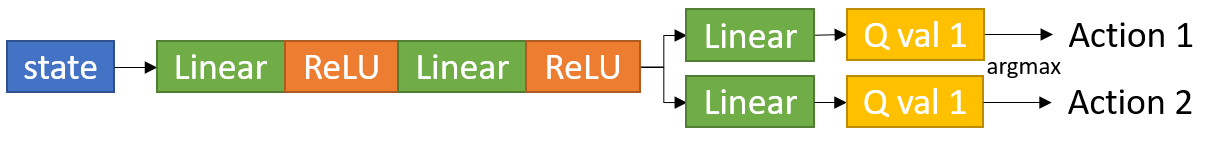}
  \caption{Structure of a Double Action DQN (2D-DQN).\label{fig:ab}}
  \vspace*{-1.0em}
\end{figure}

The reinforcement learning model is a deep Q network (DQN)
\cite{mnih2013playing} with an action branching structure 
for multi-dimensional output \cite{tavakoli2018action},
depicted in Figure \ref{fig:ab}. 
The reward structure uses knowledge of the shortest path between the
start and goal.  For each training run the reference shortest path is
obtained from the $D^*$-Lite algorithm. 
During navigation, the reward is the negative of the
distance between the agent's current position and the nearest point on
the shortest path. This reward is passed to the agent every 2 seconds
for a policy update.  The agent will also update at the end of a
training run with the following terminal reward:
\begin{equation}
	R_{goal} = \begin{cases}
		\textcolor{white}{+}1000 ({l}/{l_{min}}) & \text{if goal attained}\\
		-1000 & \text{otherwise}
	\end{cases},
\end{equation}
based on the the Success Weighted by Path Length 
\cite{anderson2018evaluation}, where $l$ and $l_{min}$ are the lengths
of the actual path taken by the robot and the reference shortest path,
respectively.

%\textcolor{red}{whose input state is the 
%sensor data} and outputs are
%the discrete $\dLA$ value and/or $\fGP$ value depending on whether the
%system is learning a single parameter or both parameters.
%When learning the two hyper-params together, we will train brand new
%networks that output two
%different values every time the agent acts, one for each
%hyper-parameter. 
%The main difference between the network we
%use and the network in \cite{tavakoli2018action} is that we use the
%naive DQN while they use the Dueling DQN. Another minor difference is
%that the experiments in \cite{tavakoli2018action} mainly use actions
%with the same action space (increasing degree of freedom) although the
%structure they propose doesn't have this limitation. In our experiments,
%the two different hyper-parameters will have different action space.

\section{Experiments and Methodology \label{sMethod}}

All of the training methods used data generated from ROS/Gazebo
simulations of robot navigation engagements in unknown environments.
The robot chosen is a Turtlebot equipped with a Microsoft Kinect sensor.
The Maze environment in Figure \ref{fig:benchmark} was the training
environment and the other environments were used for testing.  
%They are benchmark environments from \cite{SmEtAl[2020]PiPSNav}.
For pose information, the true robot pose is accessed from Gazebo.  
%\textcolor{red}{The sensor data we use are laserscan data.}

\subsection{Experimental Methodology}

\subsubsection{The Maze Environment}
To define navigation scenarios for which there is always a valid path
between any random start and goal poses, we designed a maze environment
consisting of maze walls placed within a $20 m \times 20 m$ square room.
The maze walls are composed of $0.5 m\times0.5 m$ blocks placed on a
Manhattan grid within the maze free space. Several different wall
configurations for the mazes exist. The occupancy map of these
different mazes are used as the initial global occupancy map for the
robot.

During a trial, the initial map provided for the maze will be incorrect
because of the randomized placement of obstacle cyclinders with a radius
of $0.15 m$ within the world. There are two random placement strategies,
uniform density and non-uniform density. A non-uniform placement density
is achieved by dividing the environment into $5\times5$ sub-regions and
assigning each region a random uniform density. The density
specification for a maze region is given by a minimum inter-obstacle
distance from the discrete set $\{0.75, 1.0, 1.25, 1.5\}\,m$. 

%Different local scenarios are achieved by varying the density of random
%obstacles. Here, density of obstacles means the minimum distance
%between any two obstacles. For example, 0.75m density means the
%distance between any two obstacles is larger than 0.75 meter. 
%The densities of obstacles we use are 0.75, 1.0, 1.25, and 1.5 meters. For
%density 0.75, the minimum distance between obstacles is 0.75 and all
%the maze walls are solid, which means that the robot cannot go through
%the wall; however, for higher value of density, there may be holes
%inside the maze walls, enabling the robot to take shortcuts. As the
%density value increases, there will be more holes. 
%The global costmaps only include the maze walls initially, without any information of the random obstacles or holes in the walls, and can be updated as the robot explores the world.

\subsubsection{Batch Data Collection}
To generate training data for the supervised methods, we performed a
single coordinate parameter sweep over the uniform density and robot
configuration and ran each configuration 50 times with valid random
start and goal poses lead. The result was a a total of  
$50\times5\times4=1000$ runs for $\fGP$ and 
$50\times10\times4=2000$ runs for $\dLA$. 
The robot navigation data of the robot is recorded, including the
egoTEB processed collision-space scan data and the navigation
performance data of the runs, such as whether a run succeeded,
the path length, and the runtime. 

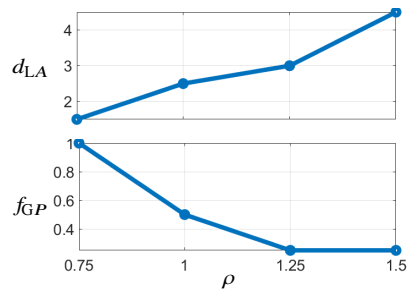
\begin{figure}[t]
  \topSpace
  \centering
  \begin{tikzpicture}[inner sep=0pt]
    % dLA vs rho
    \node[anchor=north east] (D) at (0in,0.1in)
      {{\includegraphics[scale=0.275,clip=true,trim=0.73in 0.70in
      0.5in 0.15in]{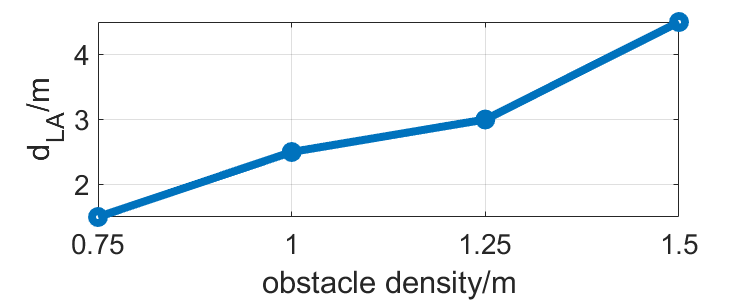}}};
    % fGP vs rho
    \node[anchor=north east,yshift=-4pt] (F) at (D.south east)
      {{\includegraphics[scale=0.275,clip=true,trim=0.50in 0.42in
      0.5in 0.15in]{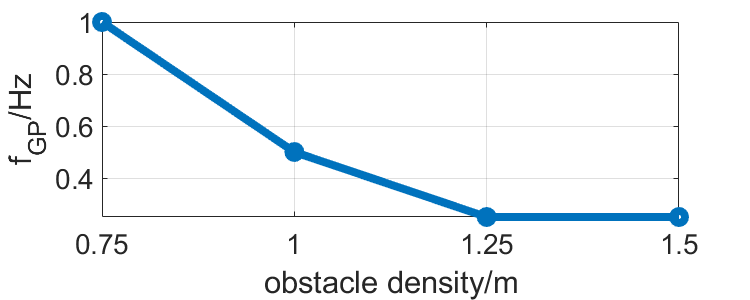}}};

    \node[anchor=north,yshift=0pt] at (F.south) {\footnotesize $\rho$};
    \node[anchor=east] (yF)  at (F.west)  {\footnotesize $\fGP$};
    \node[anchor=east]  at (D.west -| yF.east)  {\footnotesize $\dLA$};
  \end{tikzpicture}
  \caption{Best hyper-parameter versus obstacle density $\rho$.\label{fig:plot}}
  \vspace*{-1.0em}
\end{figure}

The binary success data provides a success rate for the navigation
parameter and density pairing combination.
For each density we choose the hyper-parameter value that maximizes the
success rate. The average path length is used as a tie breaker
(then the average runtime in the case of another tie).  The best value
versus the density $\rho$ defines the selection functions $\fGP(\rho)$
and $\dLA(\rho)$, which are depicted in Fig.~\ref{fig:plot}. 

%For training data generation, we run the robot in uniform density maze
%environments with different densities. 

%During each run, we will fix the hyper-parameter value, and for
%different runs, we change the hyper-parameter to different values.  

%and goal poses for each density each hyper-parameter value, resulting
%in 

%We run a series of training experiments to collect data that will be
%used to train our models, and a series of evaluation experiments where
%we test the performance of our models in unseen benchmark environments.

%The reason why we
%choose this value range for  $\dLA$ is because the sensor we
%use has a maximum range of about 5.5 meters, and the  $\dLA$
%should be larger than the minimum obstacle distance of 0.75 meter to
%identify a gap between obstacles. 

\subsubsection{Training the Neural Networks}
The classifiers were trained with cross entropy loss and the regressors
were trained with mean squared error loss, both using the SGD optimizer.
The known densities of the mazes and the best hyper-parameter values for
said densities, plus the recorded laser scan readings during the
navigation scenarios provide the training data. 

%it is possible to train a
%network with the laserscan as input and with the optimal parameter as
%the output.
%We will use the laserscan data and the chosen best hyper-parameter
%values as the ground truth data to train the supervised models. 
%Laserscan data is chosen once every 5 seconds, resulting in 89208 samples totally. 

Since training for reinforcement learning requires evaluating rollouts,
it involves a different process.  The DQN was trained in a non-uniform
environment using $\epsilon$-greedy with the reward function described in 
Section \ref{sScopeRL}, 
either from scratch or with a warm start using the collected batch data.
%In each training run, we first find the reference shortest path using 
%$D^*$-Lite algorithm. During the navigation, the agent will update once
%every 2 seconds using the distance between agent's current position and
%the shortest path as reward. The agent will also update at the end of
%navigation with success weighted by path length as reward.  

The model trained from scratch is trained for 1000 runs, and the warm
started one is trained for 800 runs after first training
with behavior cloning using the chosen best hyper-params
for 200 runs. We compare the two models to see how previous knowledge
acquired from experiments in uniform environments will influence model
performance.

\subsection{Testing and Evaluation}

The evaluation experiments are all done in non-uniform environments,
with the egoTEB planner set for dynamic reconfiguration.  Every two
seconds the parameter prediction model will predict the optimal
parameters and apply them. The baseline implementation of egoTEB uses
the fixed parameters $\dLA = 3m$ and $\fGP=1$Hz. 
%\textcolor{red}{We have experimented with modifying only one
%hyper-parameter and modifying both hyper-parameters simultaneously.}

%\subsubsection{NavTuner Learners}
%For the supervised learning method, the two supervised networks will run
%in parallel, each predicting a value for the corresponding hyper-parameter.
%For the RL method, a DQN trained to tune each hyper-parameter will
%output its value.  , and when modifying both hyper-parameters, we will train and use new models as described in \S\ref{parallel DQN}.
%
\subsubsection{Testing Environments}
Experiments are conducted in the four environments: 
maze, sector (dense), campus (dense), and office (dense).
The last three are benchmark environments from \cite{Smith2020}.
The maze environment has two sets of evaluation experiments: with the
same maze and with a maze different from the training maze.
Chaging the maze tests whether the network models have learned the
specific maze wall setup instead of the local obstacle distribution.
%The different maze is shown in Figure\ref{fig:diff}.

%The Other
%3 environments are unseen to the models, used to test the generality of
%the learned models. 
%Examples of the environments are shown in Figure \ref{fig:benchmark}. 
To vary from the training data, we generate 4 different types of random
obstacles, including 0.3m$\times$0.3m and 0.15m$\times$0.15m boxes, and
cylinders with diameters of 0.3m and 0.1m. 
The maximum number of obstacles are 200 for the maze
and sector environments, and 500 for the campus and office environments. 
In each environment, we run the 5 experiment configurations, consisting
of 0, 25\%, 50\%, 75\%, and 100\% of the maximum number of obstacles. 
%The replanning frequency is 5 Hz for local planner in all the experiments. 

\subsubsection{Evaluation Criteria}

For each run, the performance data collected is the success rate, the
path length, and the runtime. 
For the maze environment, we also run experiments where the parameters
are updated based on the best value curves (Figure \ref{fig:plot}),
using the known density.  It represents an oracle version.

\section{Results and Analysis \label{sResults}}
This section reports and analyzes the results of several experiments. 
Of the metrics recorded, the success rate exhibited the most differences
across the techniques, thus the discussion and analysis will revolve
around this quantity. The experiments performed include (a) comparative
performance of egoTEB versus end-to-end learning schemes, (b) a first
pass evaluation of outcomes for the Maze environment and a single
parameter, (c) a more complete evaluation across environments, and (d)
the extension of the best performing method to more parameters. Lastly,
there is a comparative discussion regarding contemporary works with
similar aims.

\subsection{Comparison with End-to-End Learning\label{evalE2E}}
This experiment whose outcomes are in Figure \ref{fig:benchmark} compares
the fixed parameter egoTEB to Perception-to-Decision (P2D)
\cite{PerceptionToDecision} and to the local goal classifier (LGC)
\cite{SmEtAl_Drive[2018]}. A third method, IntentionNet
\cite{intention-net} was implemented to the best of our ability
(including communication with the authors), but it would not provide
good results in the benchmark environments. Thus, it is not included in
the plots. The two methods implemented were run in an end-to-end manner,
as well as within the same hierarchical mavigation system as egoTEB.
There are 25 runs per environment. As discussed in \S\ref{NavML}, the
fixed parameter egoTEB implementation outperforms all of the
learning-based implementations. The success rate is higher, there is
a smaller gap between the max and min success rates, and the outcomes
across environments are more consistent. If end-to-end learning is to be
pursued as a navigation scheme, structurally different solutions than
those explored will be needed. For a traditionally engineered solution,
such as egoTEB, the environment sensitivity should be addressed.

\subsection{Evaluation in Maze Environments\label{evalMaze}}
Moving to the learnt policy tuners, Table \ref{tab:wall setup} contains
the success rates for the Maze environment with the maximum number of
obstacles (200) and variation of $\dLA$ only, for the same/different
maze environments.  The success rates are similar across all methods
when examining the two maze types (compare down columns); they vary by
3\% or less.  The similarity indicates
that the policy tuners are most likely learning the local obstacle
distributions and not the wall setup.  All tuner models outperform the
default values (DV), but they do not all outperform the best values
(BV). The only tuners which do are the RL implementations. 
This outcome suggests that it is beneficial to have a closed-loop,
embodied learning process whereby the policy tuner evaluates and
corrects its own performance. Doing so more effectively explores the
parameter and environment space.
%Doing so more effectively explores the parameter space.

\begin{table}[t]
  \topSpace
  \centering
  \caption{Success Rates for Maze Environments\label{tab:wall setup}}
  \addtolength{\tabcolsep}{-2pt}
  \begin{tabular}{|c|c|c|c|c|c|c|c|c|c|c|}
	\hline
	&\multirow{2}{*}{DV} &\multirow{2}{*}{BV} & \multicolumn{3}{c|}{Classifier} & \multicolumn{3}{c|}{Regressor} & \multicolumn{2}{c|}{DQN} \\\cline{4-11}
	&&& L & NN & CNN & L & NN & CNN & Sc & WS\\\hline
	Same     & 67&	78&	70&	73&	74&	69&	72&	75&	\bf 81&	80\\\hline
	Different& 68&	75&	69&	71&	74&	69&	71&	73&	\bf 80&	\bf 80\\\hline
  \end{tabular}
  \addtolength{\tabcolsep}{2pt}
  \vspace*{-1em}
\end{table}

\subsection{Evaluation in All Benchmark Environments}
The success rate outcomes across all navigation environments and for all
obstacle configuration levels are plotted in the graphs of 
Figure \ref{fig:SRvsOQ} (top). Both parameters $\fGP$ and $\dLA$ vary. The first property to note is that all
policy tuners operate at or above the success rate (SR) of the fixed
parameters egoTEB implementation. The non-negative performance impact
shows that the results from \S\ref{evalMaze} translate across obstacle
quantities and environments.
The top two performers continue to be the RL tuners since their traces
often lie above those of the others. 
Lower slopes for the policy tuners relative to that of the default
egoTEB implementation is an indication of reduced sensitivity to the
obstacle density. Several policy tuners have this property for some 
obstacle density ranges.

Figure \ref{fig:DSR} (bottom) quantifies the reduced sensitivity by
plotting bar graphs of the difference the highest and the lowest success
rates across the different obstacle count implementations.  Almost all
methods have a lower difference versus the baseline egoTEB
implementation, though some just barely. In a couple cases (mostly for
the Sector world), the higher difference is a function of a higher best
performance that increased the size of the performance gap while still
outperforming the default egoTEB. Though better by number, having a
large difference is less than ideal.  Again, the RL tuners exhibit
the best performance, with the DQN trained from scratch having reduced
the difference to 35\% (46\%) of egoTEB's on average for training from
scratch (warm start). 
One detail in favor of a warm start is that it improved the zero
obstacle success rate for the Maze and Sector cases, which is where the
default egoTEB does not have perfect performance. 
This boost is not seen for training from scratch.

%Also, the results of $\dLA$ + $\fGP$ aren't as good as those of individual hyper-param combined, which indicates that $\dLA$ and $\fGP$ are not perpendicular. But for the RL models, $\dLA$ + $\fGP$ still gives better result than individual hyper-param.

\begin{figure*}[t]
  \centering
  \begin{tikzpicture}[inner sep=0pt,outer sep=0pt]
    \node[anchor=south west] (M) at (0,0) 
      {\includegraphics[scale=0.35,clip=true,trim=0.3in 0.3in 0in 0in]{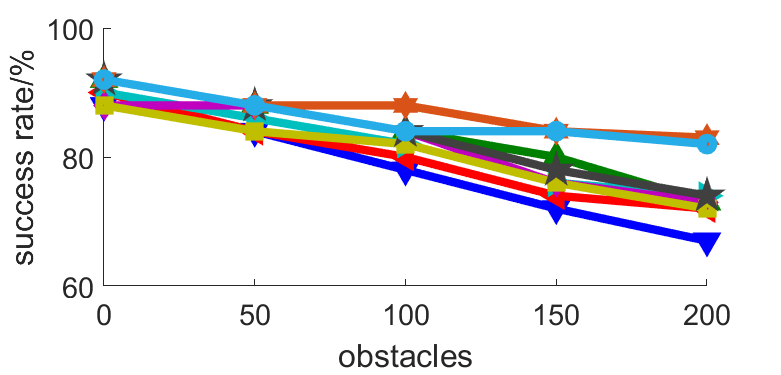}};
    \node[anchor=south west,xshift=0pt] (C) at (M.south east)
      {\includegraphics[scale=0.35,clip=true,trim=0.3in 0.3in 0in 0in]{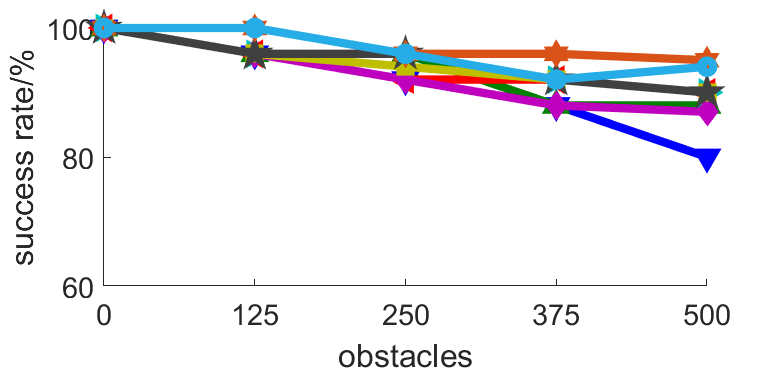}};
    \node[anchor=south west,xshift=0pt] (S) at (C.south east)
      {\includegraphics[scale=0.35,clip=true,trim=0.3in 0.3in 0in 0in]{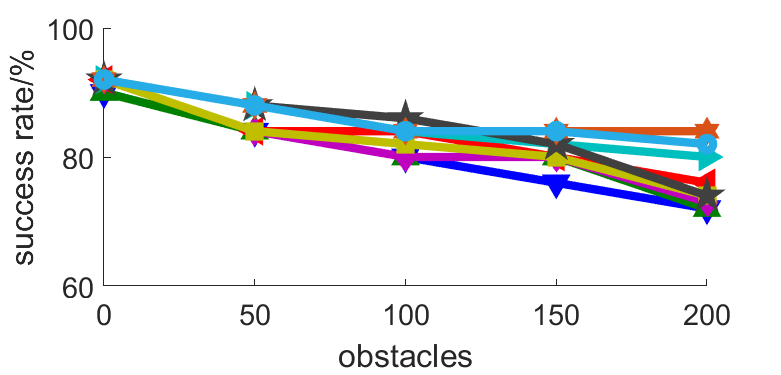}};
    \node[anchor=south west,xshift=0pt] (O) at (S.south east)
      {\includegraphics[scale=0.35,clip=true,trim=0.3in 0.3in 0in 0in]{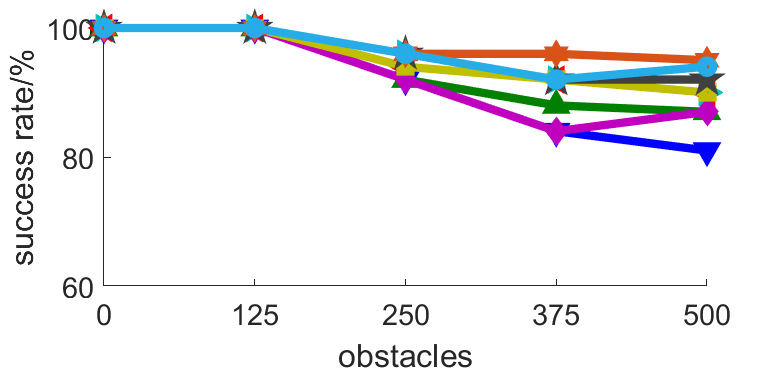}};

    \node[anchor=south, yshift=3pt] (L) at ($(C.north)!0.5!(S.north)$)
      {\includegraphics[width=1.85\columnwidth]{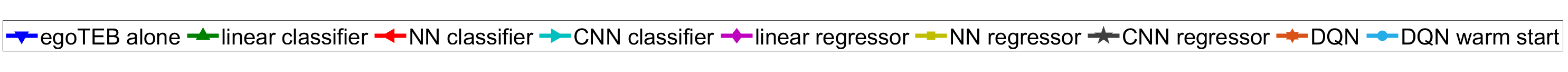}};

    % Environment labels.
    \node [anchor=south,yshift=-3pt] at (M.north) {\footnotesize Maze};
    \node [anchor=south,yshift=-3pt] at (C.north) {\footnotesize Campus};
    \node [anchor=south,yshift=-3pt] at (S.north) {\footnotesize Sector};
    \node [anchor=south,yshift=-3pt] at (O.north) {\footnotesize Office};
    \node [anchor=north,yshift=-2pt] at (M.south) 
        {\scriptsize obstacle quantity};
    \node [anchor=north,yshift=-2pt] at (C.south) 
        {\scriptsize obstacle quantity};
    \node [anchor=north,yshift=-2pt] at (S.south) 
        {\scriptsize obstacle quantity};
    \node [anchor=north,yshift=-2pt] at (O.south) 
        {\scriptsize obstacle quantity};
    \node [rotate=90,anchor=south,xshift=0.4em,yshift=2pt] at (M.west) 
        {\scriptsize SR (\%)};
  \end{tikzpicture}
  \scalebox{0.95}{
  \begin{tikzpicture}[inner sep=0pt,outer sep=0pt]
    \node (BG) at (0,0)
      {\includegraphics[width=0.95\linewidth,clip=true,trim=2.35in 0.25in 1.95in 0in]{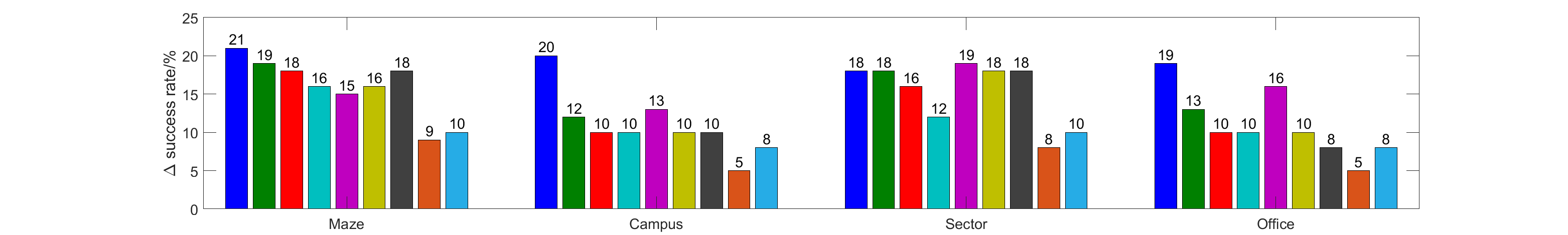}};
    \node [rotate=90,anchor=south,xshift=0em,yshift=2pt] at (BG.west)
    {\scriptsize $\Delta$SR (\%)};
    \node[anchor=north] at ($(BG.south west)+(0.91in,0in)$) 
      {\scriptsize Maze};
    \node[anchor=north] at ($(BG.south west)+(2.57in,0in)$) 
      {\scriptsize Campus};
    \node[anchor=north] at ($(BG.south west)+(4.22in,0in)$) 
      {\scriptsize Sector};
    \node[anchor=north] at ($(BG.south west)+(5.875in,0in)$) 
      {\scriptsize Office};
  \end{tikzpicture}}
  \caption{Top: Success Rate vs Obstacle Quantity plots for the different
    environments.\label{fig:SRvsOQ}
    Bottom: Difference in Success Rate between best and worst cases for
    the different environments.\label{fig:DSR} Legend color coding
    applies to both figures.}
    \vspace*{-1.5em}
\end{figure*}
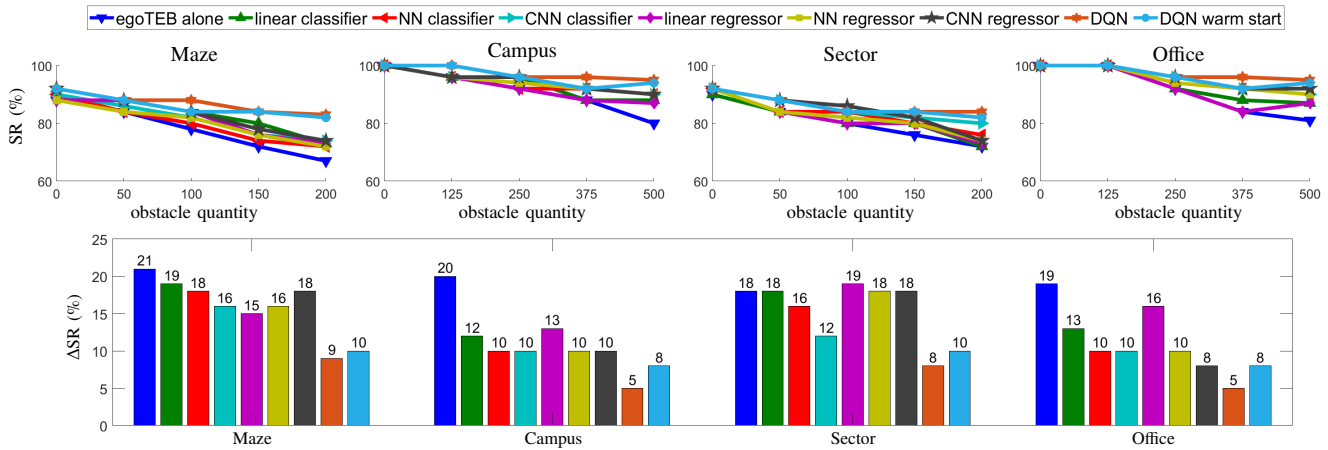
\begin{figure}[t]
  %\vspace*{0.5em}
  \begin{tikzpicture}[inner sep=0pt, outer sep=0pt]
    \node (BG) at (0,0)
      {\includegraphics[width=0.9\columnwidth,clip=true,trim=1.05in 0in
      1.5in 0in]{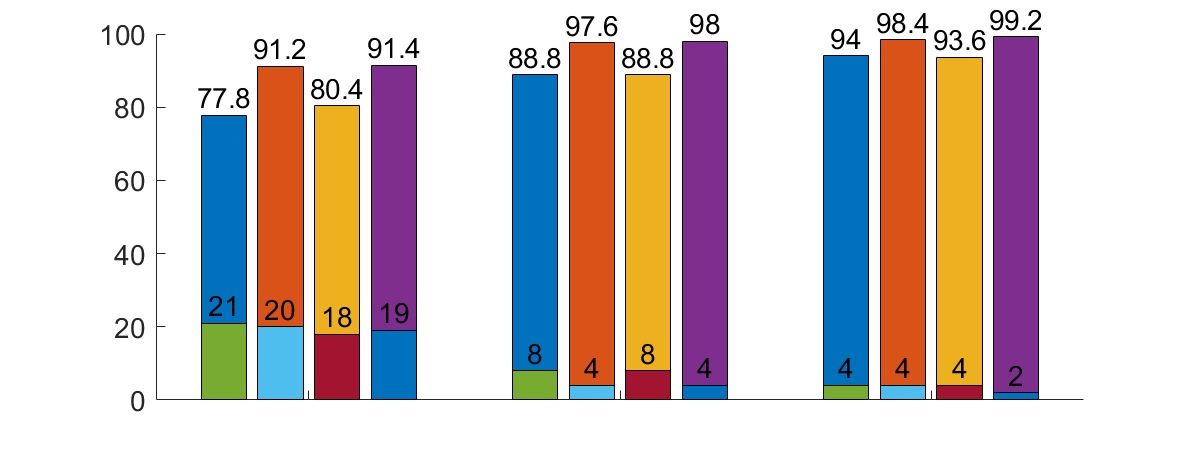}};

    \node[anchor=south,rotate=90] at (BG.west) 
      {\scriptsize SR / $\Delta SR$ (\%)};
    \node[anchor=north, xshift=0.675in] at (BG.south west) 
      {\scriptsize egoTEB};
    \node[anchor=north, xshift=1.65in] at (BG.south west) 
      {\scriptsize 2D NavTuner};
    \node[anchor=north, xshift=2.635in] at (BG.south west) 
      {\scriptsize 7D NavTuner};

    \node[anchor=north, xshift=28.75pt,yshift=0.85em] (eM) at (BG.south west) 
      {\scriptsize M};
    \node[anchor=north west, xshift=13.5pt] (eC) at (eM.north west) 
      {\scriptsize C};
    \node[anchor=north west, xshift=13.5pt] (eS) at (eC.north west) 
      {\scriptsize S};
    \node[anchor=north west, xshift=13.5pt] (eO) at (eS.north west) 
      {\scriptsize O};

    \node[anchor=north, xshift=101.0pt,yshift=0.85em] (tM) at (BG.south west) 
      {\scriptsize M};
    \node[anchor=north west, xshift=13.5] (tC) at (tM.north west) 
      {\scriptsize C};
    \node[anchor=north west, xshift=13.5] (tS) at (tC.north west) 
      {\scriptsize S};
    \node[anchor=north west, xshift=13.5] (tO) at (tS.north west) 
      {\scriptsize O};

    \node[anchor=north, xshift=173.5pt,yshift=0.85em] (sM) at (BG.south west) 
      {\scriptsize M};
    \node[anchor=north west, xshift=13.5pt] (sC) at (sM.north west) 
      {\scriptsize C};
    \node[anchor=north west, xshift=13.5pt] (sS) at (sC.north west) 
      {\scriptsize S};
    \node[anchor=north west, xshift=13.5pt] (sO) at (sS.north west) 
      {\scriptsize O};

  \end{tikzpicture}
  \caption{Success Rate and Difference in Success Rate bar plots for
  different environments and navigation systems.
  The taller bars are the success rate with the actual number printed
  above them.  The shorter overlaid bars are the difference in success
  rate values with the actual difference printed above them.%
  \label{compNavTuner}}
  \vspace*{-1.0em}
\end{figure}
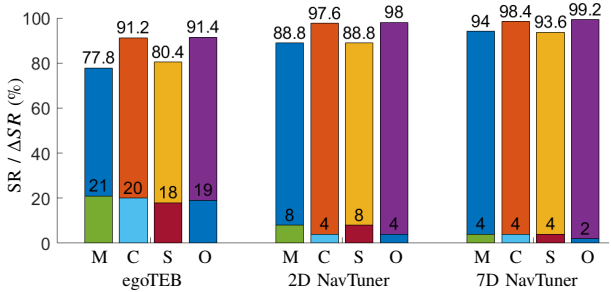

\subsection{Evaluation of a 7D-DQN NavTuner}

Since the RL-based policy tuner had the best performance, we will denote
any such implementation to be a {\em NavTuner} for short-hand. Given the
good performance of the 2D NavTuner, a 7D version with a 7D-DQN was
deployed and trained using the same training process but with 4000
rollouts (taking 4.2 days to train a model).  The 2D-DQN was 
trained again from scratch for the same 4000 rollouts.
The global planner parameter value to output was $\fGP$, and the
egoTEB local planner parameter values to output were
$\dLA$, {\em selection cost hysteresis}, path {\em switching blocking
period}, {\em selection prefers initial plan}, 
{\em inflation distance}, and the {\em number of poses in the
feasibility check} (integer). 
Comparative results between egoTEB, the 2D-DQN, and the 7D-DQN are in
the bar charts of Figure \ref{compNavTuner} with the actual numerical
values printed about the bars.  The 7D NavTuner boosts the performance
and reduces the difference, as seen by the higher bars and the lower
difference values.  The zero obstacle performance boost not seen for the
2D NavTuner case with training from scratch, now occurs. The campus and
office worlds have near perfect success rates across all implementations
as evidence by the high success rate and low difference. Recall that
they start at 100\% with no obstacles. Compared to egoTEB, the average
success rate (marginalized across all environments) for egoTEB with
a 7D-DQN NavTuner improves by 13\%, from 85.2\% to 96.3\%. The
sensitivity drops from an average difference of 19.5\% for egoTEB to
3.5\% for the 7D NavTuner version (82\% lower). When compared to the 2D
NavTuner, the sensitivity drop is from 6\% to 3.5\% (42\% lower).

\subsection{Discussion and Comparison to Other Tuning Models}

The online tuning of navigation parameters using deep learning methods
is relatively recent. This section discusses the outcomes of a few
methods in relation to the NavTuner findings. The approach in 
\cite{bhardwaj2020differentiable} aimed to arrive at a differentiable
model for the deep network training process to correct for structural
deficiencies in soft-constraint optimal control solvers.  While the
method did lower sensitivity to obstacle configurations, it did not
improve the success rate.  It is doubtful that such an approach could
work for the class of soft-constraint optimization solvers studied.
Typically, soft constraints are addressed using a scale-space method
that solves the problem multiple times under different soft-constraint
parameters to incrementally approach the hard-constraint solution.
Even then feasibility is not guaranteed.  The egoTEB algorithm builds on
TEB \cite{TEB}, which also employs a soft-constraint optimal control
solver using factor graphs like \cite{bhardwaj2020differentiable}. 
By solving a highly localized problem for detected gaps, many of the
deficiencies of soft-constraint solvers are avoided. Especially since
multiple solutions are obtained and feasibility checking is performed.
That NavTuner boosts performance and reduces sensitivity using a
model-free and non-differentiable method suggests that trying to derive
a solution like \cite{bhardwaj2020differentiable} may not be necessary
when there are learning schemes that do not require it.  Model-free RL
will be more effective when the parameters are difficult to
differentiate, as holds for the egoTEB navigation parameters.

\begin{figure*}[t]
  \begin{tikzpicture}[inner sep=0pt, outer sep=0pt]
    \node[anchor=north west] (M) at (0in,0in)
      {\includegraphics[width=1.7in]{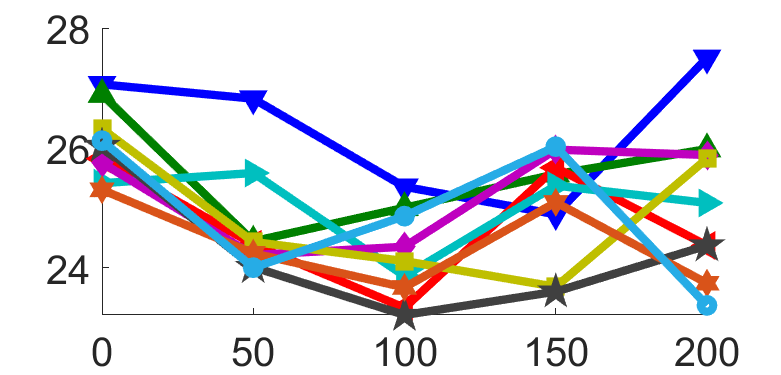}};
    \node[anchor=north west] (C) at (M.north east)
      {\includegraphics[width=1.7in]{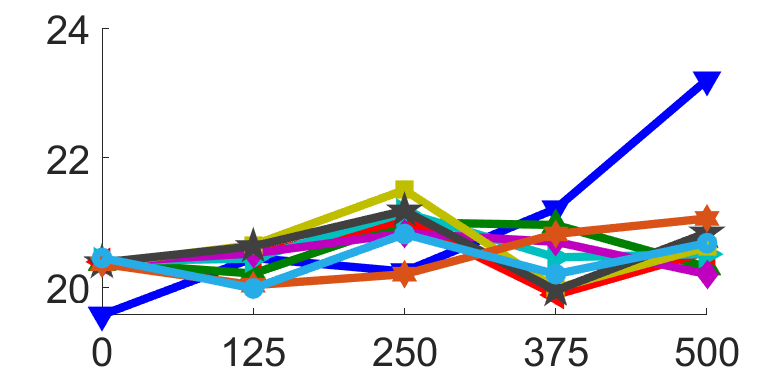}};
    \node[anchor=north west] (S) at (C.north east)
 {\includegraphics[width=1.7in]{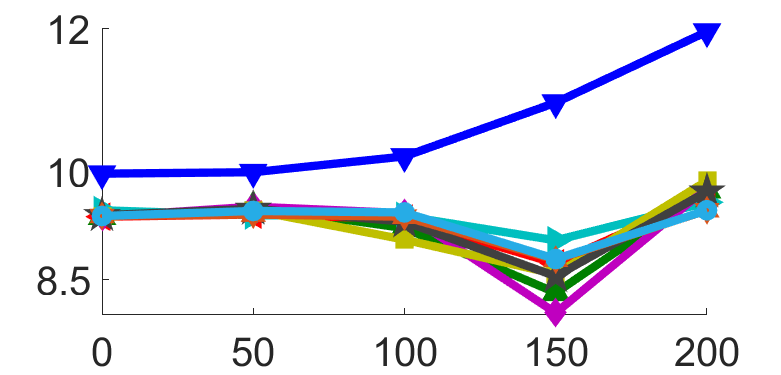}};
    \node[anchor=north west] (O) at (S.north east)
 {\includegraphics[width=1.7in]{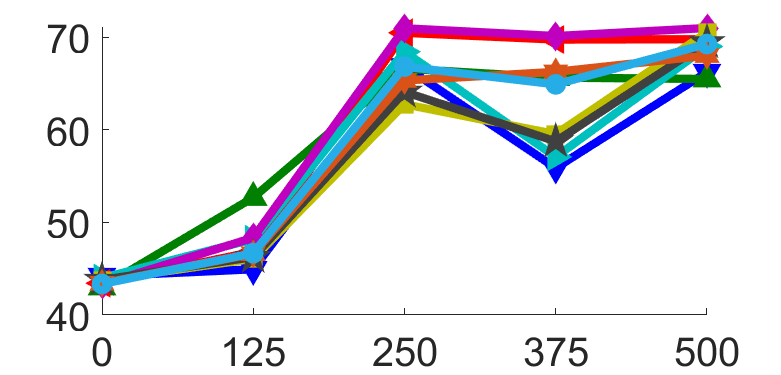}};

    \node[rotate=90,xshift=6pt,yshift=1pt,anchor=south] at (M.west) {\scriptsize path length (m)};
    \node[anchor=north,xshift=-2em] at (M.north) {\scriptsize Maze};
    \node[anchor=north,xshift=-2em] at (C.north) {\scriptsize Campus};
    \node[anchor=north,xshift=-2em] at (S.north) {\scriptsize Sector};
    \node[anchor=north,xshift=-2em] at (O.north) {\scriptsize Office};

    \node[anchor=north west,yshift=-5pt] (MI) at (M.south west)
        {\includegraphics[width=1.7in]{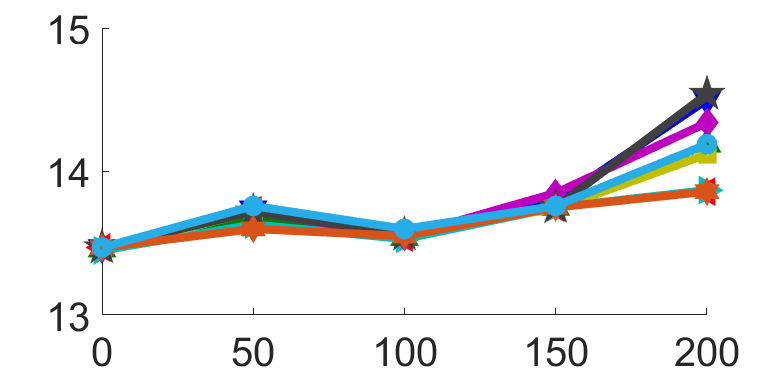}};
    \node[anchor=north west] (CI) at (MI.north east)
        {\includegraphics[width=1.7in]{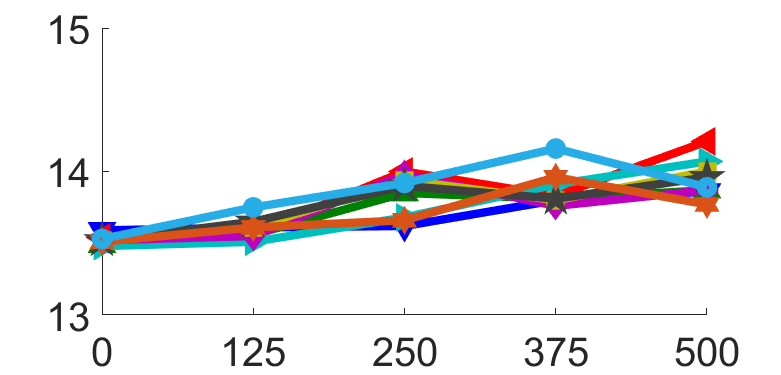}};
    \node[anchor=north west] (SI) at (CI.north east)
        {\includegraphics[width=1.7in]{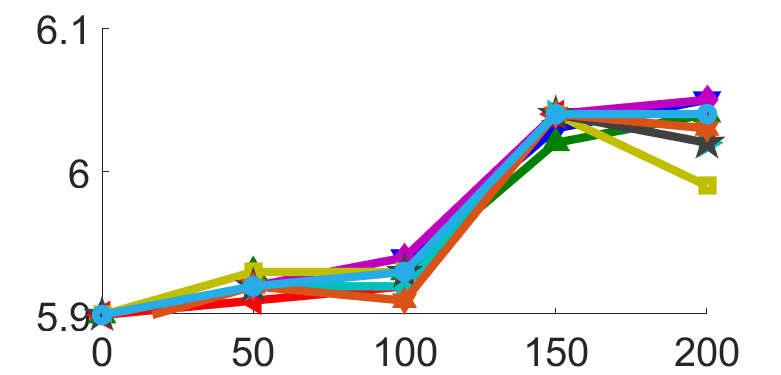}};
    \node[anchor=north west] (OI) at (SI.north east)
        {\includegraphics[width=1.7in]{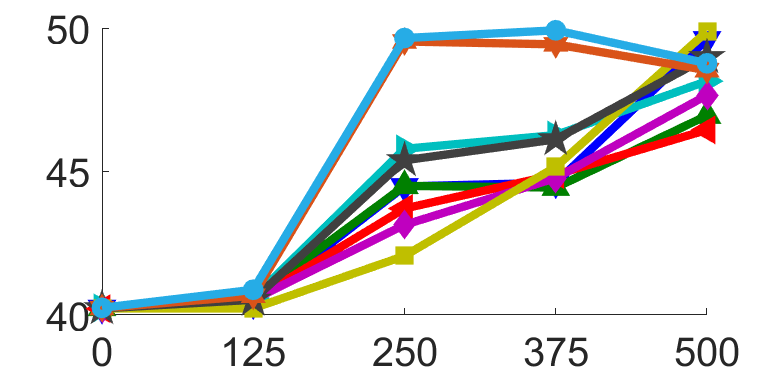}};

    \node[rotate=90,xshift=6pt,yshift=1pt,anchor=south] at (MI.west) {\scriptsize path length (m)};
    \node[anchor=north,xshift=1em] at (MI.south) {\scriptsize \# obstacles};
    \node[anchor=north,xshift=1em] at (CI.south) {\scriptsize \# obstacles};
    \node[anchor=north,xshift=1em] at (SI.south) {\scriptsize \# obstacles};
    \node[anchor=north,xshift=1em] at (OI.south) {\scriptsize \# obstacles};

  \end{tikzpicture}
  \caption{{\bf Top row:} Path length comparison across the different
  approaches based on successful paths for each approach. {\bf Bottom
  row:} Path length comparison restricted to the common successful path
  runs. Legend same as in Figure 5.\label{fig:LenComp}}
%\end{figure*}
%\begin{figure*}[t]
\vspace*{0.5em}
  \begin{tikzpicture}[inner sep=0pt, outer sep=0pt]
    \node[anchor=north west] (M) at (0in,0in)
        {\includegraphics[width=1.7in]{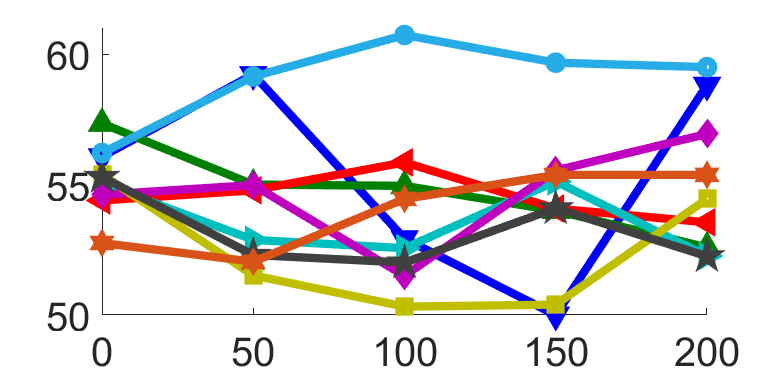}};
    \node[anchor=north west] (C) at (M.north east)
        {\includegraphics[width=1.7in]{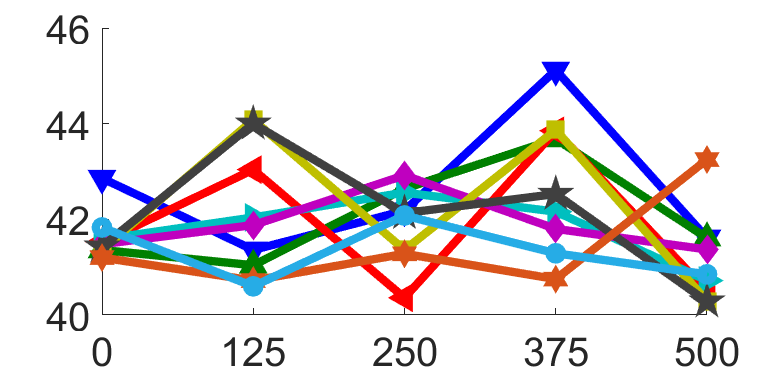}};
    \node[anchor=north west] (S) at (C.north east)
        {\includegraphics[width=1.7in]{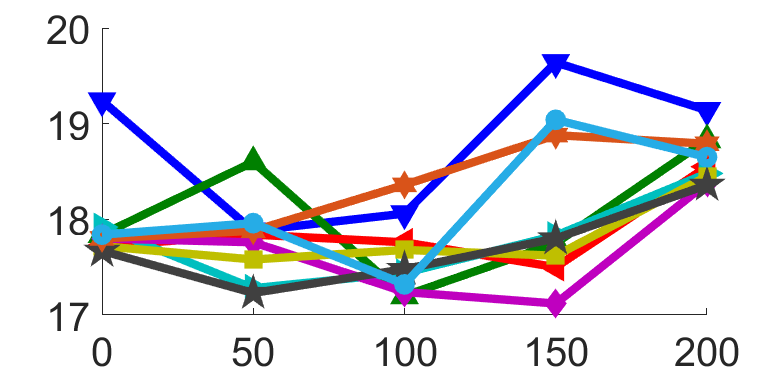}};
    \node[anchor=north west] (O) at (S.north east)
        {\includegraphics[width=1.7in]{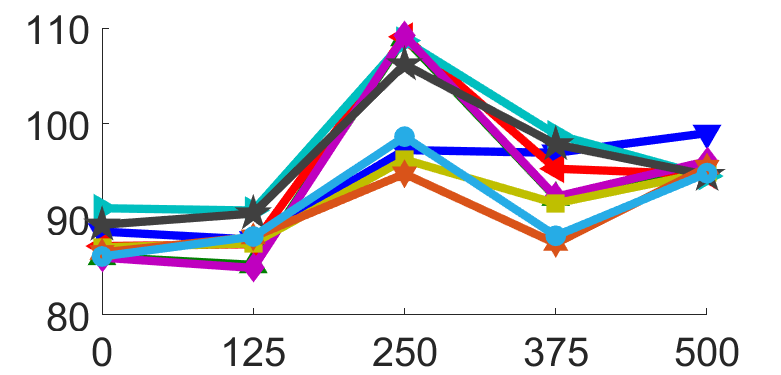}};

    \node[rotate=90,xshift=6pt,yshift=1pt,anchor=south] at (M.west) {\scriptsize time (s)};
    \node[anchor=north,xshift=-2em] at (M.north) {\scriptsize Maze};
    \node[anchor=north,xshift=-2em] at (C.north) {\scriptsize Campus};
    \node[anchor=north,xshift=-2em] at (S.north) {\scriptsize Sector};
    \node[anchor=north,xshift=-2em] at (O.north) {\scriptsize Office};

    \node[anchor=north west,yshift=-5pt] (MI) at (M.south west)
      {\includegraphics[width=1.7in]{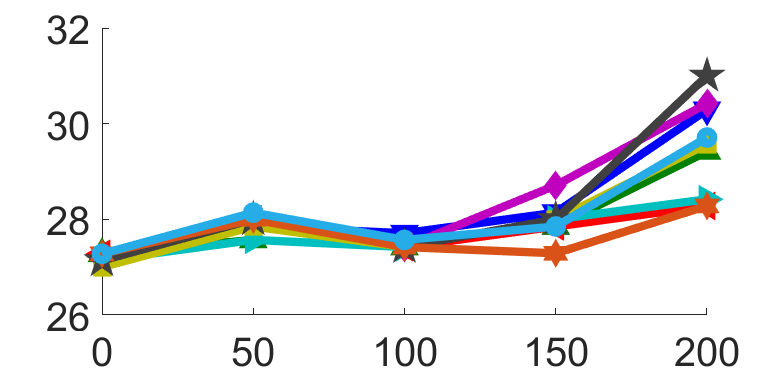}};
    \node[anchor=north west] (CI) at (MI.north east)
      {\includegraphics[width=1.7in]{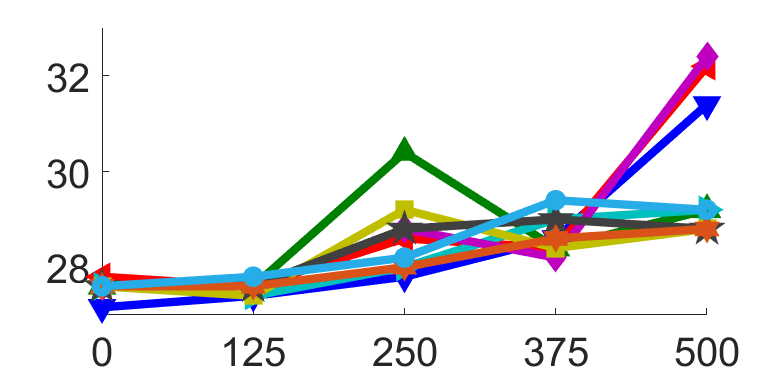}};
    \node[anchor=north west] (SI) at (CI.north east)
      {\includegraphics[width=1.7in]{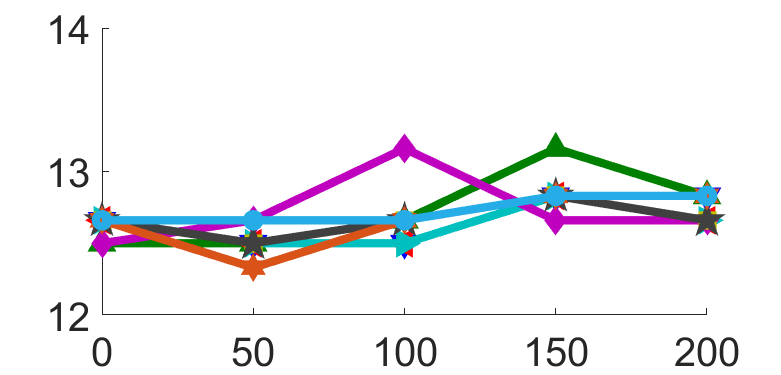}};
    \node[anchor=north west] (OI) at (SI.north east)
      {\includegraphics[width=1.7in]{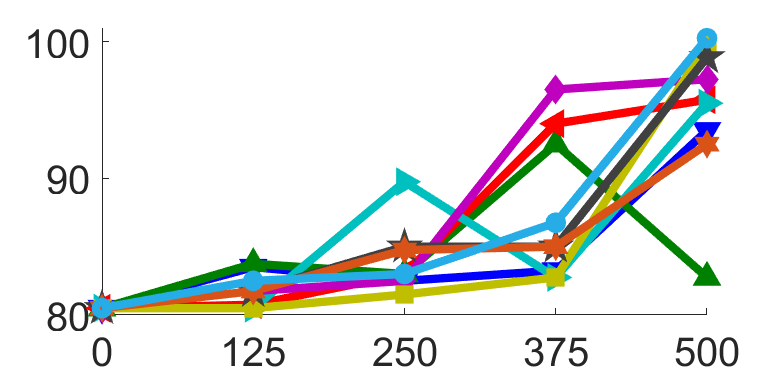}};

    \node[rotate=90,xshift=6pt,yshift=1pt,anchor=south] at (MI.west)
    {\scriptsize time (s)};

  \end{tikzpicture}
  \caption{{\bf Top row:} Traversal time comparison across the different
  approaches based on successful paths for each approach. {\bf Bottom
  row:} Traversal time comparison restricted to the common successful
  path runs. Legend same as in Figure 5.\label{fig:TimeComp}}
  \vspace*{-0.5em}
\end{figure*}

The approaches in \cite{wang2020appli,xiao2020appld,xu2020applr} (APPLI,
APPLD, and APPLR) are the closest to NavTuner.  For APPLI, rather than report
success rate a time penalty was assigned to the run, and only the
average completion times were reported. APPLI reported a 43\% relative
improvement compared to fixed parameters DWA \cite{dynamicwindow}, 
but also exhibited worse performance for a smaller percentage of trials
versus DWA. 
APPLD discussed navigation failures in the text. Under APPLD, the
designed maze was traversed every time, while DWA failed for the great
majority. APPLD requires human demonstration and cannot leverage
simulation to automatically learn a policy tuner. Hence the
implementation of APPLR based on RL, much like NavTuner. Evaluation of
APPLR versus the DWA version was mixed, in the sense that APPLR performed
better for a subset of experiments, but worse or equivalently for
another (with these latter being the harder scenarios).  APPLR had high
variance in traversal time versus DWA, which led to the mixed outcomes.
%In contrast NavTuner demonstrated a boost in performance across the board.
Though we did record path length and runtime as performance metrics,
we found that they did not capture performance as well the success rate did.  
Furthermore, it is more difficult to compare when the approaches do not
have comparable success rates, which may explain the mixed outcomes of
the APPLx implementations.  For example, Figures \ref{fig:LenComp} and
\ref{fig:TimeComp} compare the path lengths and traversal times,
respectively. There are two rows, with the top being the averages across
all successful runs for each method and the bottom being the average for
the common successful runs across all obstacle configurations, thus it
compares comparable navigation scenarios and outcomes vs methods. 
Performance is comparable when looking at the runs for which all
succeeded (bottom rows). Divergence of statistics happens when including
the differing outcomes, usually in the form of more variance or a shift
in performance. The increased path length for Office by the RL methods
in the bottom row may seem incorrect, but it persists in the top row and
these methods have the best success rate. This shift most likely
indicates that a longer, circuitous path is being taken, but that is
more favorable. Being longer is not disadvantageous. For the Sector case
(top row), egoTEB has a monotonic increase while the others do not.  The
common dip suggests that they are also leveraging some of the global
scene structure that egoTEB with fixed parameters cannot. Comparing path
length and traversal time has more nuance than just needing to be
shorter. 

%All of
%the solutions tested exhibited similar values \textcolor{red}{MORE
%DISCUSSION}, thus they were not included in the presentation. To see
%these outcomes, please view the {\em Supplementary Material} in the
%project repository \cite{ivagit_NavTuner}.

%\textcolor{red}{In contrast NavTuner had demonstrably improved success
%rates and lowered sensitivity to obstacle configurations and
%environments.}
One potential difference is that the APPLx navigation scenarios
resembled tunnel like environments solvable using a forward motion
wander navigation method, which might be the cause of the mixed
outcomes.  Additionally, the sensor was a high or full field of view
laser scanner, while NavTuner deployed a limited field-of-view depth
imaging sensor leading to the possibility of passing near unsensed
regions.  NavTuner learns to modulate the parameters in a way that these
partially observed situations do not significantly impact navigation
performance.  No sensitivity analysis for the APPLx algorithms
illuminated clear differences in outcomes for a variable environmental
parameter, thus it is more difficult to ascertain the overall benefits
of APPLR over DWA beyond traversal time (the text does note more
recovery behavior state events, but does not quantify them).  In the
sensitivity analysis here, the one variable that exhibited the most
sensitivity, and also happens to be the most important, is the success
rate.  Using RL, NavTuner 
%learns to modulate navigation
%parameters in response to the local environment such that performance
%and robustness improve. 
effectively learns a family of scene-sensitive navigation policies to
select from during navigation, thereby leading to performance and
robustness improvements.

\section{Conclusion \label{sConc}}
This study provided experimental evidence using controllable simulations
of navigation scenarios to show that online adaptive tuning of
navigation parameters for engineered navigation systems can improve
run-time performance and sensitivity to environmental conditions. The
sensitivity associated to traditionally designed systems is often used
as a justification for replacing them with deep learning models. 
This paper started by showing that contemporary deep learning navigation
methods could not match the performance of a fixed parameter navigation
system.  Furthermore, it showed that learning a family of navigation
policies for online navigation parameter tuning further boosts
performance. The improvements are more substantive in comparison to
existing methods with similar contexts, suggesting that those studies were
incomplete. The code is open-sourced \cite{ivagit_NavTuner}.

The use of the free space measurements as the input data should permit
some level of sim-to-real transfer since it is an intermediate
representation. Deep learning on intermediate representations, or with
them, is less sensitive to source signal noise since it has been
processed out. The APPLx family of tuners provides evidence that
transfer can happen.  Future work aims to confirm this conjecture for
NavTuner. 
%Furthermore, the same concept can be applied to other engineered systems
%with parameter sensitivity, such as SLAM systems.

%There is still room for future exploration. It still needs exploring
%whether the models can generalize well to real world environments where
%both the sensor data and the robot actuation may be noisy. Also, we can
%try to tune more hyper-parameters and see whether there will be further
%performance boost.

%Also, the local planner is replanning at a fixed frequency, which means that the robot will receive a new path from the planner when the robot has gone along only a small fraction of the old one. This may mitigate the effect of different $\dLA$. 
%Apart from these, we currently choose the best value of lookahead distance based on success rate and path length. It may be better if we design a heuristic function based on the combination of success rate, path length, runtime, or other performance data. Based on the distribution of the DQN outputs, we suspect that the best lookahead distance and the best global planning frequency may depend on the time for the agent to reach the obstacle or the horizon instead of the distance between the agent and the obstacle. This may be validated by conducting similar experiments with robots of different maximum speed and compare the results.

%\section*{Acknowledgment}
%
%The preferred spelling of the word ``acknowledgment'' in America is without 

%\balance
\bibliographystyle{IEEEtran}
\bibliography{reference,planning3}

\begin{thebibliography}{10}
\providecommand{\url}[1]{#1}
\csname url@rmstyle\endcsname
\providecommand{\newblock}{\relax}
\providecommand{\bibinfo}[2]{#2}
\providecommand\BIBentrySTDinterwordspacing{\spaceskip=0pt\relax}
\providecommand\BIBentryALTinterwordstretchfactor{4}
\providecommand\BIBentryALTinterwordspacing{\spaceskip=\fontdimen2\font plus
\BIBentryALTinterwordstretchfactor\fontdimen3\font minus
  \fontdimen4\font\relax}
\providecommand\BIBforeignlanguage[2]{{%
\expandafter\ifx\csname l@#1\endcsname\relax
\typeout{** WARNING: IEEEtran.bst: No hyphenation pattern has been}%
\typeout{** loaded for the language `#1'. Using the pattern for}%
\typeout{** the default language instead.}%
\else
\language=\csname l@#1\endcsname
\fi
#2}}

\bibitem{TEB}
C.~R{\"o}smann, W.~Feiten, T.~W{\"o}sch, F.~Hoffmann, and T.~Bertram,
  ``{Efficient Trajectory Optimization Using a Sparse Model},'' in
  \emph{{European Conference on Mobile Robots}}, Sept 2013, pp. 138--143.

\bibitem{smith2020egoteb}
J.~S. Smith, R.~Xu, and P.~Vela, ``egoteb: Egocentric, perception space
  navigation using timed-elastic-bands,'' in \emph{IEEE International
  Conference on Robotics and Automation}.\hskip 1em plus 0.5em minus
  0.4em\relax IEEE, 2020, pp. 2703--2709.

\bibitem{SmEtAl_Drive[2018]}
\BIBentryALTinterwordspacing
J.~Smith, J.~Hwang, F.~Chu, and P.~Vela, ``Learning to navigate: Exploiting
  deep networks to inform sample-based planning during vision-based
  navigation,'' \emph{arxiv/CoRR}, vol. abs/1801.05132, 2018. [Online].
  Available: \url{http://arxiv.org/abs/1801.05132}
\BIBentrySTDinterwordspacing

\bibitem{PerceptionToDecision}
M.~{Pfeiffer}, M.~{Schaeuble}, J.~{Nieto}, R.~{Siegwart}, and C.~{Cadena},
  ``From perception to decision: A data-driven approach to end-to-end motion
  planning for autonomous ground robots,'' in \emph{2017 IEEE International
  Conference on Robotics and Automation (ICRA)}, May 2017, pp. 1527--1533.

\bibitem{intention-net}
\BIBentryALTinterwordspacing
W.~Gao, D.~Hsu, W.~S. Lee, S.~Shen, and K.~Subramanian, ``{Intention-Net:
  Integrating Planning and Deep Learning for Goal-Directed Autonomous
  Navigation},'' \emph{CoRR}, vol. abs/1710.05627, 2017. [Online]. Available:
  \url{http://arxiv.org/abs/1710.05627;
  http://dblp.org/rec/bib/journals/corr/abs-1710-05627}
\BIBentrySTDinterwordspacing

\bibitem{XiEtAl[2020]NavLearnSurvey}
X.~Xiao, B.~Liu, G.~Warnell, and P.~Stone, ``Motion control for mobile robot
  navigation using machine learning: a survey,'' \emph{arXiv}.

\bibitem{PRM-RL}
\BIBentryALTinterwordspacing
A.~Faust, O.~Ramirez, M.~Fiser, K.~Oslund, A.~Francis, J.~Davidson, and
  L.~Tapia, ``Prm-rl: Long-range robotic navigation tasks by combining
  reinforcement learning and sampling-based planning,'' Brisbane, Australia,
  2018, pp. 5113--5120. [Online]. Available:
  \url{https://arxiv.org/abs/1710.03937}
\BIBentrySTDinterwordspacing

\bibitem{Kim2016}
\BIBentryALTinterwordspacing
B.~Kim and J.~Pineau, ``Socially adaptive path planning in human environments
  using inverse reinforcement learning,'' \emph{International Journal of Social
  Robotics}, vol.~8, no.~1, pp. 51--66, Jan 2016. [Online]. Available:
  \url{https://doi.org/10.1007/s12369-015-0310-2}
\BIBentrySTDinterwordspacing

\bibitem{chaplot2020learning}
D.~S. Chaplot, D.~Gandhi, S.~Gupta, A.~Gupta, and R.~Salakhutdinov, ``Learning
  to explore using active neural slam,'' \emph{arXiv preprint
  arXiv:2004.05155}, 2020.

\bibitem{bansal2020combining}
S.~Bansal, V.~Tolani, S.~Gupta, J.~Malik, and C.~Tomlin, ``Combining optimal
  control and learning for visual navigation in novel environments,'' in
  \emph{Conference on Robot Learning}.\hskip 1em plus 0.5em minus 0.4em\relax
  PMLR, 2020, pp. 420--429.

\bibitem{self-supervised-rl-gen-graphs}
G.~{Kahn}, A.~{Villaflor}, B.~{Ding}, P.~{Abbeel}, and S.~{Levine},
  ``Self-supervised deep reinforcement learning with generalized computation
  graphs for robot navigation,'' in \emph{2018 IEEE International Conference on
  Robotics and Automation (ICRA)}, May 2018, pp. 1--8.

\bibitem{target-driven-visual-rl}
Y.~{Zhu}, R.~{Mottaghi}, E.~{Kolve}, J.~J. {Lim}, A.~{Gupta}, L.~{Fei-Fei}, and
  A.~{Farhadi}, ``Target-driven visual navigation in indoor scenes using deep
  reinforcement learning,'' in \emph{2017 IEEE International Conference on
  Robotics and Automation (ICRA)}, May 2017, pp. 3357--3364.

\bibitem{bruce2017one}
J.~Bruce, N.~S{\"u}nderhauf, P.~Mirowski, R.~Hadsell, and M.~Milford,
  ``One-shot reinforcement learning for robot navigation with interactive
  replay,'' \emph{arXiv preprint arXiv:1711.10137}, 2017.

\bibitem{DeepVisuoMotor}
\BIBentryALTinterwordspacing
S.~Levine, C.~Finn, T.~Darrell, and P.~Abbeel, ``End-to-end training of deep
  visuomotor policies,'' \emph{J. Mach. Learn. Res.}, vol.~17, no.~1, pp.
  1334--1373, Jan. 2016. [Online]. Available:
  \url{http://dl.acm.org/citation.cfm?id=2946645.2946684}
\BIBentrySTDinterwordspacing

\bibitem{Choi2011}
\BIBentryALTinterwordspacing
J.-M. Choi, S.-J. Lee, and M.~Won, ``Self-learning navigation algorithm for
  vision-based mobile robots using machine learning algorithms,'' \emph{Journal
  of Mechanical Science and Technology}, vol.~25, no.~1, pp. 247--254, Jan
  2011. [Online]. Available: \url{https://doi.org/10.1007/s12206-010-1023-y}
\BIBentrySTDinterwordspacing

\bibitem{virtual-to-real-rl}
L.~{Tai}, G.~{Paolo}, and M.~{Liu}, ``Virtual-to-real deep reinforcement
  learning: Continuous control of mobile robots for mapless navigation,'' in
  \emph{2017 IEEE/RSJ International Conference on Intelligent Robots and
  Systems (IROS)}, Sep. 2017, pp. 31--36.

\bibitem{drl-successor-features}
J.~{Zhang}, J.~T. {Springenberg}, J.~{Boedecker}, and W.~{Burgard}, ``Deep
  reinforcement learning with successor features for navigation across similar
  environments,'' in \emph{2017 IEEE/RSJ International Conference on
  Intelligent Robots and Systems (IROS)}, Sep. 2017, pp. 2371--2378.

\bibitem{nav-aux-tasks}
\BIBentryALTinterwordspacing
P.~Mirowski, R.~Pascanu, F.~Viola, H.~Soyer, A.~J. Ballard, A.~Banino,
  M.~Denil, R.~Goroshin, L.~Sifre, K.~Kavukcuoglu, D.~Kumaran, and R.~Hadsell,
  ``Learning to navigate in complex environments,'' \emph{CoRR}, vol.
  abs/1611.03673, 2016. [Online]. Available:
  \url{http://arxiv.org/abs/1611.03673}
\BIBentrySTDinterwordspacing

\bibitem{ye2020auxiliary}
J.~Ye, D.~Batra, E.~Wijmans, and A.~Das, ``Auxiliary tasks speed up learning
  pointgoal navigation,'' \emph{arXiv preprint arXiv:2007.04561}, 2020.

\bibitem{sax2019learning}
A.~Sax, J.~O. Zhang, B.~Emi, A.~Zamir, S.~Savarese, L.~Guibas, and J.~Malik,
  ``Learning to navigate using mid-level visual priors,'' \emph{arXiv preprint
  arXiv:1912.11121}, 2019.

\bibitem{kumar2020learning}
A.~Kumar, S.~Gupta, and J.~Malik, ``Learning navigation subroutines from
  egocentric videos,'' in \emph{Conference on Robot Learning}, 2020, pp.
  617--626.

\bibitem{gao2017intention}
W.~Gao, D.~Hsu, W.~S. Lee, S.~Shen, and K.~Subramanian, ``Intention-net:
  Integrating planning and deep learning for goal-directed autonomous
  navigation,'' \emph{arXiv preprint arXiv:1710.05627}, 2017.

\bibitem{8326229}
Y.~{Kim}, J.~{Jang}, and S.~{Yun}, ``End-to-end deep learning for autonomous
  navigation of mobile robot,'' in \emph{2018 IEEE International Conference on
  Consumer Electronics (ICCE)}, Jan 2018, pp. 1--6.

\bibitem{dynamicwindow}
D.~Fox, W.~Burgard, and S.~Thrun, ``{The dynamic window approach to collision
  avoidance},'' \emph{Robotics Automation Magazine, IEEE}, vol.~4, no.~1, pp.
  23--33, Mar 1997.

\bibitem{teb-first}
C.~{Roesmann}, W.~{Feiten}, T.~{Woesch}, F.~{Hoffmann}, and T.~{Bertram},
  ``Trajectory modification considering dynamic constraints of autonomous
  robots,'' in \emph{ROBOTIK 2012; 7th German Conference on Robotics}, May
  2012, pp. 1--6.

\bibitem{uneven-unstructured-indoor}
C.~{Wang}, L.~{Meng}, S.~{She}, I.~M. {Mitchell}, T.~{Li}, F.~{Tung}, W.~{Wan},
  M.~Q.~. {Meng}, and C.~W. {de Silva}, ``Autonomous mobile robot navigation in
  uneven and unstructured indoor environments,'' in \emph{2017 IEEE/RSJ
  International Conference on Intelligent Robots and Systems (IROS)}, Sep.
  2017, pp. 109--116.

\bibitem{feurer2019hyperparameter}
M.~Feurer and F.~Hutter, ``Hyperparameter optimization,'' in \emph{Automated
  Machine Learning}.\hskip 1em plus 0.5em minus 0.4em\relax Springer, Cham,
  2019, pp. 3--33.

\bibitem{yu2020hyper}
T.~Yu and H.~Zhu, ``Hyper-parameter optimization: A review of algorithms and
  applications,'' \emph{arXiv preprint arXiv:2003.05689}, 2020.

\bibitem{huang2019automatic}
C.~{Huang}, B.~{Yuan}, Y.~{Li}, and X.~{Yao}, ``Automatic parameter tuning
  using bayesian optimization method,'' in \emph{2019 IEEE Congress on
  Evolutionary Computation (CEC)}, 2019, pp. 2090--2097.

\bibitem{AutoRL}
H.~L. {Chiang}, A.~{Faust}, M.~{Fiser}, and A.~{Francis}, ``Learning navigation
  behaviors end-to-end with autorl,'' \emph{IEEE Robotics and Automation
  Letters}, vol.~4, no.~2, pp. 2007--2014, April 2019.

\bibitem{cano2018automatic}
J.~Cano, Y.~Yang, B.~Bodin, V.~Nagarajan, and M.~O'Boyle, ``Automatic parameter
  tuning of motion planning algorithms,'' in \emph{2018 IEEE/RSJ International
  Conference on Intelligent Robots and Systems (IROS)}.\hskip 1em plus 0.5em
  minus 0.4em\relax IEEE, 2018, pp. 8103--8109.

\bibitem{burger2017automated}
R.~Burger, M.~Bharatheesha, M.~van Eert, and R.~Babu{\v{s}}ka, ``Automated
  tuning and configuration of path planning algorithms,'' in \emph{2017 IEEE
  International Conference on Robotics and Automation (ICRA)}.\hskip 1em plus
  0.5em minus 0.4em\relax IEEE, 2017, pp. 4371--4376.

\bibitem{berkenkamp2016bayesian}
F.~Berkenkamp, A.~Krause, and A.~P. Schoellig, ``Bayesian optimization with
  safety constraints: safe and automatic parameter tuning in robotics,''
  \emph{arXiv preprint arXiv:1602.04450}, 2016.

\bibitem{bhardwaj2020differentiable}
M.~Bhardwaj, B.~Boots, and M.~Mukadam, ``Differentiable gaussian process motion
  planning,'' in \emph{2020 IEEE International Conference on Robotics and
  Automation (ICRA)}.\hskip 1em plus 0.5em minus 0.4em\relax IEEE, 2020, pp.
  10\,598--10\,604.

\bibitem{wang2020appli}
Z.~Wang, X.~Xiao, B.~Liu, G.~Warnell, and P.~Stone, ``Appli: Adaptive planner
  parameter learning from interventions,'' \emph{arXiv preprint
  arXiv:2011.00400}, 2020.

\bibitem{xiao2020appld}
X.~Xiao, B.~Liu, G.~Warnell, J.~Fink, and P.~Stone, ``Appld: Adaptive planner
  parameter learning from demonstration,'' \emph{arXiv preprint
  arXiv:2004.00116}, 2020.

\bibitem{xu2020applr}
Z.~Xu, G.~Dhamankar, A.~Nair, X.~Xiao, G.~Warnell, B.~Liu, Z.~Wang, and
  P.~Stone, ``Applr: Adaptive planner parameter learning from reinforcement,''
  \emph{arXiv preprint arXiv:2011.00397}, 2020.

\bibitem{Smith2020}
\BIBentryALTinterwordspacing
J.~S. Smith, S.~Feng, F.~Lyu, and P.~A. Vela, \emph{Real-Time Egocentric
  Navigation Using 3D Sensing}.\hskip 1em plus 0.5em minus 0.4em\relax Cham:
  Springer International Publishing, 2020, pp. 431--484. [Online]. Available:
  \url{https://doi.org/10.1007/978-3-030-22587-2_14}
\BIBentrySTDinterwordspacing

\bibitem{mnih2013playing}
V.~Mnih, K.~Kavukcuoglu, D.~Silver, A.~Graves, I.~Antonoglou, D.~Wierstra, and
  M.~Riedmiller, ``Playing atari with deep reinforcement learning,''
  \emph{arXiv preprint arXiv:1312.5602}, 2013.

\bibitem{tavakoli2018action}
A.~Tavakoli, F.~Pardo, and P.~Kormushev, ``Action branching architectures for
  deep reinforcement learning,'' in \emph{Thirty-Second AAAI Conference on
  Artificial Intelligence}, 2018.

\bibitem{anderson2018evaluation}
P.~Anderson, A.~Chang, D.~S. Chaplot, A.~Dosovitskiy, S.~Gupta, V.~Koltun,
  J.~Kosecka, J.~Malik, R.~Mottaghi, M.~Savva, \emph{et~al.}, ``On evaluation
  of embodied navigation agents,'' \emph{arXiv preprint arXiv:1807.06757},
  2018.

\bibitem{quigley2009ros}
M.~Quigley, K.~Conley, B.~Gerkey, J.~Faust, T.~Foote, J.~Leibs, R.~Wheeler, and
  A.~Y. Ng, ``Ros: an open-source robot operating system,'' in \emph{ICRA
  workshop on open source software}, vol.~3, no. 3.2.\hskip 1em plus 0.5em
  minus 0.4em\relax Kobe, Japan, 2009, p.~5.

\bibitem{ivagit_NavTuner}
\BIBentryALTinterwordspacing
H.~Ma, J.~Smith, and P.~Vela, ``Ivalab: Navtune git repository,'' 2021.
  [Online]. Available: \url{\url{https://github.com/ivaROS/NavTuner}}
\BIBentrySTDinterwordspacing

\end{thebibliography}

%\input{dump.tex}

%\newpage
%\newpage
%\section{MOVED TO END}
%\input{appendix.tex}
\end{document}